%% file: main.tex
\documentclass[journal]{IEEEtran}

\usepackage[margin=1in]{geometry}
\usepackage{array}
\usepackage{booktabs}

\usepackage{graphicx}
\usepackage{multirow}
\usepackage{makecell}

\usepackage{graphicx}
\usepackage{adjustbox}
\usepackage{xcolor}
\usepackage{amsmath,amssymb,amsfonts}
\usepackage{bm}
\usepackage{url}

\usepackage{tikz}
\usepackage{overpic}

\begin{document}
\title{Restricted Receptive Fields for Face
Verification}

\author{Kagan Ozturk, Aman Bhatta, Haiyu Wu, Patrick Flynn, Fellow, IEEE\\ and Kevin W. Bowyer, Life Fellow, IEEE
\thanks{The authors are with the Computer Science and Engineering Department, University of Notre Dame, Notre Dame, IN, 46556, USA (e-mail: kztrk@nd.edu).}}%

\maketitle

\input{sec/0_abstract}


\IEEEpeerreviewmaketitle
\input{sec/1_intro}

\input{sec/2_rel_work}

\input{sec/3_method}

\input{sec/4_results}

\input{sec/5_quantative_analysis}
\input{sec/6_conclusion}

\section*{Acknowledgment}
Kagan Ozturk is partly supported by the Ministry of National Education of Türkiye.

\ifCLASSOPTIONcaptionsoff
  \newpage
\fi

\bibliographystyle{IEEEtran}
\bibliography{IEEEabrv, main}

\begin{IEEEbiography}[{\includegraphics[width=1in,height=1.25in,clip,keepaspectratio]{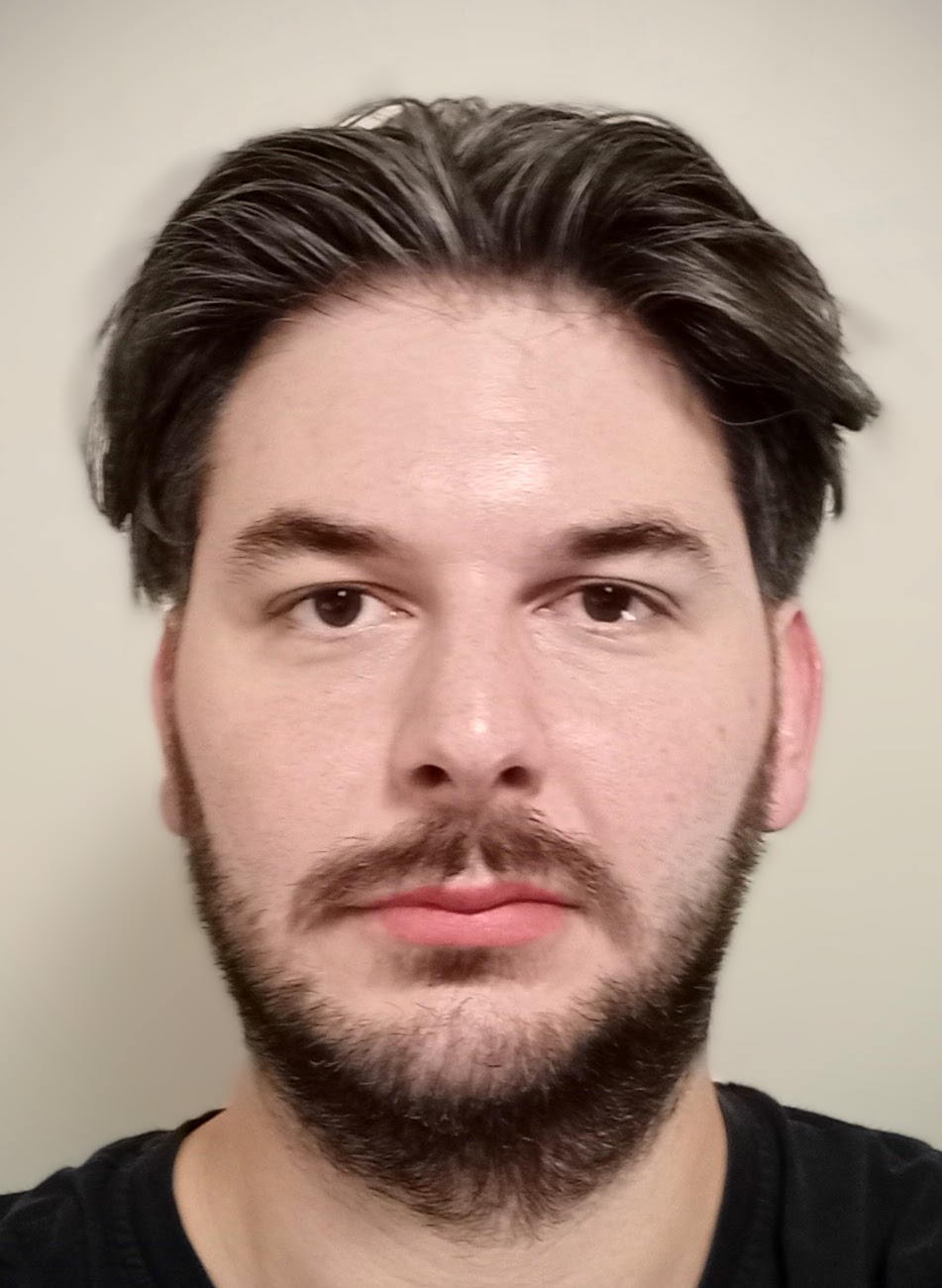}}]{Kagan Ozturk} received the M.S. degree in Computer Engineering from the Akdeniz University. He is currently pursuing the Ph.D. degree in Computer Science and Engineering at the University of Notre Dame. He is partly supported by the Ministry of National Education of Türkiye for his Ph.D. studies. His research interests include computer vision, machine learning and biometrics.
\end{IEEEbiography}
\vfill

\begin{IEEEbiography}[{\includegraphics[width=1in,height=1.25in,clip,keepaspectratio]{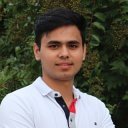}}]{Aman Bhatta} received his B.S. degree in Mechanical Engineering from the University of Mississippi in 2021 and his Ph.D. degree in Computer Science and Engineering from the University of Notre Dame in 2025. He is currently working as a Machine Learning Scientist at Wayfair. His research interests include multi-modal learning, computer vision, and machine learning.
\end{IEEEbiography}

\vfill

\begin{IEEEbiography}[{\includegraphics[width=1in,height=1.25in,clip,keepaspectratio]{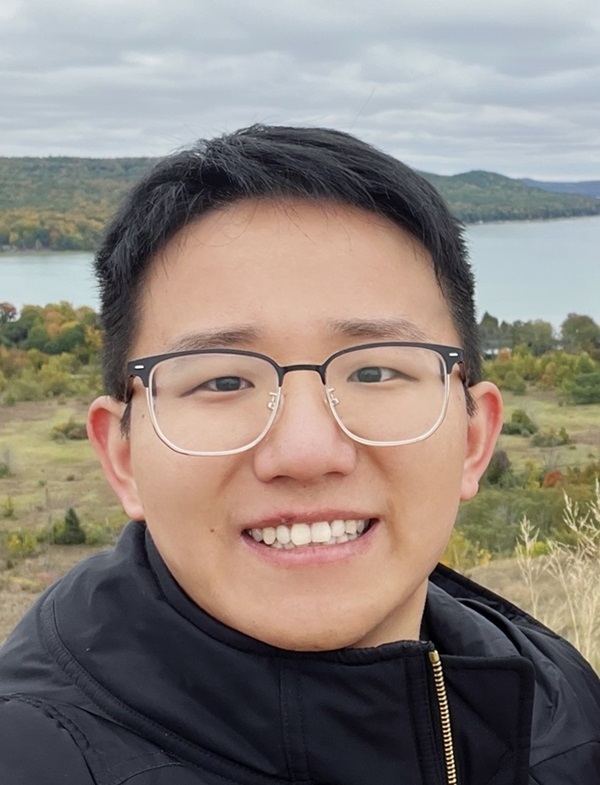}}]{Haiyu Wu} received his B.S. degree in Electrical Engineering from Northern Arizona University in 2020 and his Ph.D. degree in Computer Science and Engineering from the University of Notre Dame in 2025. He is currently working as a Research Scientist at Altos Labs. His research interests include multi-modal foundation model, world model, and AI for biology.
\end{IEEEbiography}
\vfill

\begin{IEEEbiography}[{\includegraphics[width=1in,height=1.25in,clip,keepaspectratio]{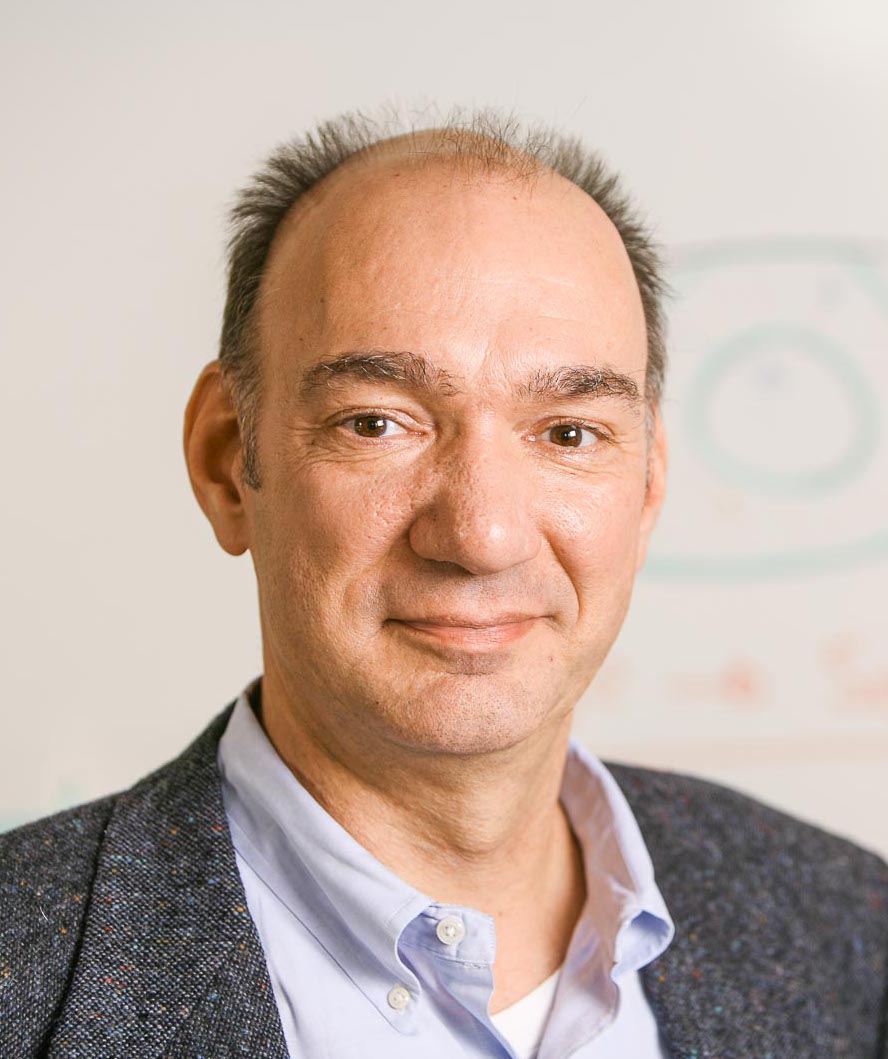}}]{Kevin W. Bowyer (Life Fellow, IEEE)} is the Schubmehl-Prein Family Professor of Computer Science and Engineering at the University of Notre Dame.  Professor Bowyer is a Fellow of the American Academy for the Advancement of Science “for distinguished contributions to the field of computer vision and pattern recognition, biometrics, object recognition and data science”, a Fellow of the IEEE “for contributions to algorithms for recognizing objects in images”, and a Fellow of the IAPR “for contributions to computer vision, pattern recognition and biometrics”.  He received a Technical Achievement Award from the IEEE Computer Society, and both the Meritorious Service Award and the Leadership Award from the IEEE Biometrics Council.  Professor Bowyer served as Editor-In-Chief of both the IEEE Transactions on Biometrics, Behavior, and Identity Science and the IEEE Transactions on Pattern Analysis and Machine Intelligence.  
\end{IEEEbiography}
\vfill

\begin{IEEEbiography}[{\includegraphics[width=1in,height=1.25in,clip,keepaspectratio]{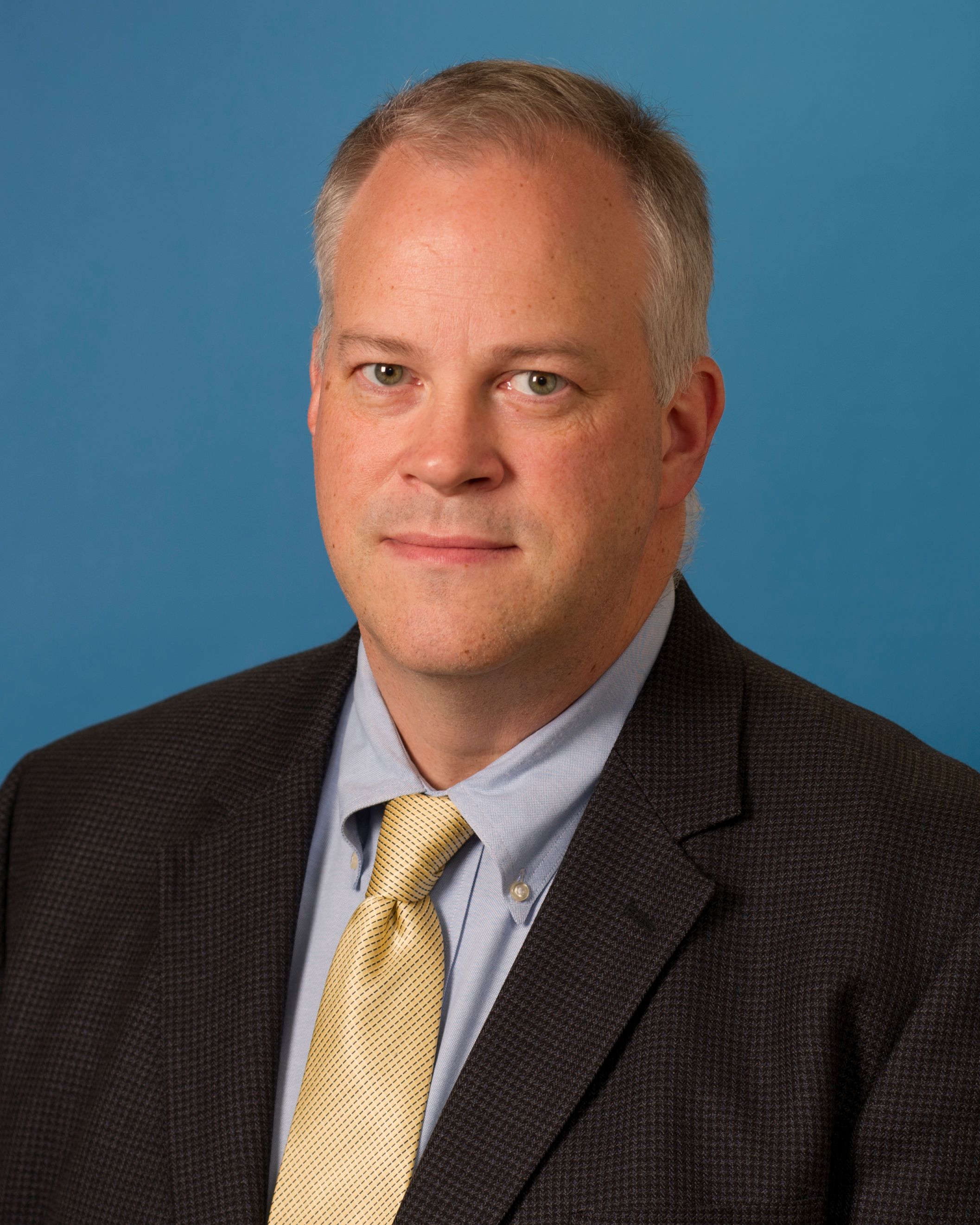}}]{Patrick J. Flynn (Fellow, IEEE)} is Fritz Duda Family Professor of Engineering and Professor of Computer Science and Engineering at the University of Notre Dame.  He received the Ph.D. in Computer Science in 1990 from Michigan State University. He has held faculty positions at Notre Dame, Washington State University, and The Ohio State University.  His research interests include computer vision, biometrics, and image processing. Dr. Flynn is a past Editor-in-Chief of the IEEE Biometrics Compendium, a past Associate Editor-in-Chief of IEEE Transactions on Pattern Analysis and Machine Intelligence, and a past Associate Editor of IEEE Transactions on Image Processing and IEEE Transactions on Information Forensics and Security.
\end{IEEEbiography}

\vfill

\end{document}

%% file: sec/0_abstract.tex
\begin{abstract}

Understanding how deep neural networks make decisions is crucial for analyzing their behavior and diagnosing failure cases. In computer vision, a common approach to improve interpretability is to assign importance to individual pixels using post-hoc methods. Although they are widely used to explain black-box models, their fidelity to the model’s actual reasoning is uncertain due to the lack of reliable evaluation metrics. This limitation motivates an alternative approach, which is to design models whose decision processes are inherently interpretable. To this end, we propose a face similarity metric that breaks down global similarity into contributions from restricted receptive fields. Our method defines the similarity between two face images as the sum of patch-level similarity scores, providing a locally additive explanation without relying on post-hoc analysis. We show that the proposed approach achieves competitive verification performance even with patches as small as $28 \times 28$ within $112 \times 112$ face images, and surpasses state-of-the-art methods when using $56 \times 56$ patches.

\end{abstract}

%% file: sec/1_intro.tex
\section{Introduction}
\label{sec:intro}

\IEEEPARstart{E}{xplainable} AI approaches have been extensively employed to analyze the decision-making processes of vision models \cite{buhrmester2021analysis, dovsilovic2018explainable, samek2019towards, adadi2018peeking, arrieta2020explainable, dwivedi2023explainable, vilone2021notions, ali2023explainable}. These approaches often generate heatmaps that visualize pixel-level contributions through a post-hoc analysis. Although such visualizations offer qualitative insights into model behavior, quantitative evaluation remains challenging, and the reliability of these explanations has been questioned in several studies \cite{laugel2019dangers, ghorbani2019interpretation, zhou2022feature, zhou2021evaluating, rudin2019stop}.

\input{figures/intro}

In face recognition, convolutional neural network (CNN) representations have been effectively leveraged to improve verification performance \cite{kim2022adaface, deng2019arcface, parkhi2015deep, wang2018cosface, zhao2003face, liu2017sphereface, wang2018additive}. Although achieving low error rates on unseen test samples enhances model credibility, the complexity of learning high-level representations directly from raw pixels makes it difficult to understand the model’s decision-making process. Many recent works attempt to interpret these models using post-hoc approaches, which introduce additional computational overhead after the decision is made and typically lack quantitative justification of the generated explanations \cite{mery2022black, huber2024efficient, knoche2023explainable, lu2024towards, huber2024recognition}.

As opposed to post-hoc approaches, interpretable models emphasize constrained design choices to make decisions inherently more interpretable \cite{rudin2019stop, rudin2022interpretable, agarwal2021neural, molnar2020interpretable, murdoch2019definitions, linardatos2020explainable}. One prominent example is ProtoPNet~\cite{chen2019looks}, which represents model outputs as a weighted sum of similarities between the input and prototypes learned from the training images. This formulation enables localized and case-based reasoning by relating regions of the input image to the prototypes. However, its applicability is limited to closed-set recognition, as the prototypes are drawn from training data.

In this work, we introduce RRFNet to optimize face recognition through the fusion of patch-level similarity scores. Unlike the traditional approach of learning a single feature vector for a similarity assessment, RRFNet defines global similarity as the sum of local similarities (illustrated in Fig. \ref{fig:intro}). Since this formulation decomposes similarity into contributions from sub-regions, it yields saliency maps as an inherent part of the decision process, eliminating the need for computationally expensive post-hoc approaches.

We evaluate our approach using two patch sizes, $28 \times 28$ and $56 \times 56$, on $112 \times 112$ face images. Although restricted receptive fields constrain the available spatial context, we surprisingly find that this design choice improves verification accuracy compared to state-of-the-art methods when $56 \times 56$ patches are used for similarity measurement. Even with patches as small as $28 \times 28$, our method achieves competitive verification performance. Note, performing verification based on local similarities inherently provides regional explanations, leading to more interpretable decision-making compared to global feature extraction approaches.

This paper is organized as follows. In Section \ref{sec:relatedwork}, we review the differences between post-hoc explanation methods and inherently interpretable approaches. Section \ref{sec:method} introduces two approaches for measuring similarity from patch representations. First, in Section \ref{sec:region-based}, we define a global distance metric formulated as a weighted sum of region-based similarities. Next, in Section \ref{sec:rrfnet}, we present RRFNet and show that with a small modification to the ResNet architecture, similarity decisions become more interpretable in terms of patch-level similarities. Experimental results for both methods are reported in Section \ref{experiment}, followed by quantitative analyses on explainability in Section \ref{sec:quantative}. Finally, Section \ref{sec:conclusion} presents conclusions and discusses directions for future research.

%% file: figures/intro.tex
\begin{figure}[t]
    \raggedright
    \centering
    \includegraphics[width=0.99\linewidth]{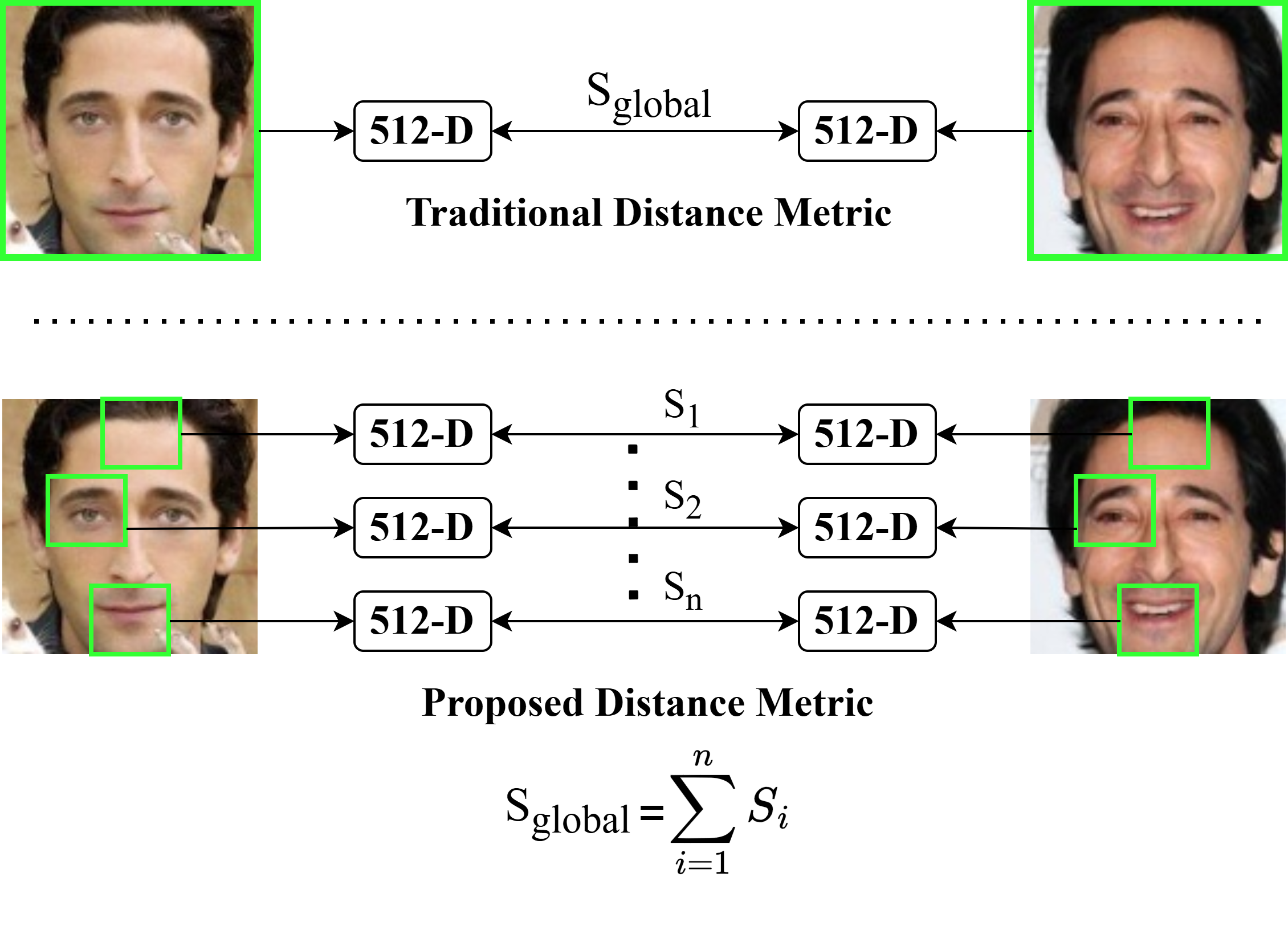}
    \vspace{-1em}
    \caption{
Comparison of the traditional (top) and the proposed approaches (bottom). In the traditional approach, face similarity is measured using a single global representation. Because feature extraction relies on black-box models, the resulting similarity score offers no insight into the decision process. In contrast, our approach extracts representations from restricted receptive fields and computes the overall similarity score as the sum of local similarities, enhancing human understanding through patch-level decomposition.}
    \label{fig:intro}
\end{figure}

%% file: sec/2_rel_work.tex
\section{Related Work}
\label{sec:relatedwork}

\input{figures/patches}

\input{figures/heatmap}

Recent advances in representation learning have enabled solutions to complex, high-dimensional problems, with applications in high-stakes areas such as security and healthcare \cite{rudin2019stop, zytek2021sibyl, gao2019towards, stiglic2020interpretability}. The unprecedented accuracy rates of these models enhances their credibility, but their inherent complexity makes their decision-making processes difficult for humans to interpret. It has been shown that slight modifications at pixel level, that are not noticeable to humans, can dramatically change predictions \cite{akhtar2018threat, andriushchenko2020square, narodytska2017simple, su2019one, yuan2019adversarial}. Given the widespread deployment of these systems, there is an urgent need for the current black-box models to become more interpretable in order to promote trust.

Interpretability and explainability of recent computer vision models have been discussed in many works \cite{linardatos2020explainable, rudin2019stop, rudin2022interpretable, molnar2020interpretable, murdoch2019definitions}. Since these two terms can have domain specific goals, differentiating them with strict definitions is challenging. One commonly accepted view of the distinction between them is as follows. Explanation methods aim to show the importance of pixels in the decision process after the black-box model is trained. 
On the other hand, interpretable methods aim to constrain predictive models so that their reasoning processes are more understandable to humans \cite{rudin2022interpretable}. While some works try to increase the interpretability through training a second model via knowledge distillation after the model is trained, most works try to modify the black-box model during training, aiming to create more transparent decisions \cite{li2022egnn, liu2018improving, li2020tnt, li2024hybrid, chen2019looks, brendel2018approximating}.

A great amount of research has focused on post-hoc explanation approaches, where the decision of a black-box model is explained by generating heatmaps that highlight important pixels that influence the model's predictions most. These visual explanations can help identify relevant regions, offering insights into where the model looks when making decisions. However, evaluating the quality of these explanations remains an unsolved challenge. Furthermore, post-hoc methods can be misleading, as similar heatmaps are often generated for both correct and incorrect predictions, undermining the reliability of these explanations \cite{tan2023considerations, aivodji2019fairwashing, ribeiro2016should, adebayo2018sanity, nie2018theoretical, laugel2019dangers}. While it has been shown that searching sub-regions can increase faithfulness of the explanations \cite{chen2024less, chen2025less, chen2025interpreting}, computational overhead limits the practical deployment.   

Another line of research in recent years focuses on incorporating interpretability directly into the vision models \cite{wang2023learning, chen2019looks, donnelly2022deformable, carmichael2024pixel, xcos}. The main component of these models is to decompose images into smaller parts and make predictions based on a weighted combination of these parts. In \cite{chen2019looks}, they employ a prototype learning approach to explain the decision through part-based evidence. Prototypes are assigned to each class based on representative image patches from the training set, and the final prediction is calculated as a weighted sum of similarity to these prototypes. While this part-based evidence framework enhances interpretability, it is restricted to classes present in the training set, limiting its generalization to unseen categories.

Recent works in face recognition have explored the use of post-hoc explanation approaches \cite{mery2022black, huber2024efficient, knoche2023explainable, lu2024towards, chen2023sim2word, eberle2020building}. In contrast, our work aims to make decisions inherently interpretable by leveraging local similarities. Similar to our approach, a self-interpretable face recognition model in \cite{xcos} is trained to provide patch-level similarity scores alongside its predictions. They introduce an xCos module to highlight the model’s regional attention, while the network’s receptive field still covers the entire image. In contrast, our method explicitly restricts the receptive field, ensuring that network decisions are directly determined by patch-level similarities. Furthermore, we quantitatively demonstrate that our approach performs significantly better on faithfulness metrics (see Section \ref{sec:quantative}).

\textbf{Patch-level Feature Learning.}  While global feature learning approaches have shown great success in recent years \cite{krizhevsky2012imagenet, he2016deep}, BagNet~\cite{brendel2018approximating} investigates the use of patch representations for object recognition. It restricts the receptive field of a CNN to enable local feature learning, while it results in a significant drop in accuracy. In face recognition, patch-level representations were extensively studied prior to the deep learning era~\cite{ahonen2004face, brunelli1993face, heisele2003face, wright2008robust, martinez2002recognizing, liao2012partial, patchbased, ZHANG201828}. The typical approach involves extracting local features using hand-crafted methods and then fusing information from different regions. For instance, \cite{patchbased} combines the outputs of nearest neighbor classifiers trained on different face patches. However, it uses the same set of subjects for both training and testing, lacking an evaluation on unseen subjects. Similarly, while \cite{ZHANG201828} uses hierarchical patch relationships to enhance local matching, its performance evaluation remains limited to fixed gallery subjects. In contrast to the traditional local feature extraction strategies, recent methods
have shifted toward global feature learning~\cite{deng2019arcface, kim2022adaface, uniface, schroff2015facenet, parkhi2015deep} to achieve state-of-the-art verification performance. In this work, we demonstrate that feature learning does not need to operate at a global level to achieve high accuracy rates. Unlike traditional fusion methods, RRFNet integrates local feature learning with score fusion, demonstrating performance competitive with state-of-the-art methods.

%% file: figures/patches.tex
\begin{figure*}[tb]
    \centering
    \begin{adjustbox}{max width=\textwidth}
    \begin{tabular}{c@{\hskip 1cm}c@{\hskip 0.7cm}c}
        \begin{minipage}{0.4\textwidth}
            \centering
            \setlength{\tabcolsep}{1.5pt}
            \begin{tabular}{ccccccc}
                \begin{tabular}{c} 
                \includegraphics[width=0.13\textwidth]{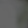} \\[-4pt] {\footnotesize (0,0)} \end{tabular} &
                \begin{tabular}{c} \includegraphics[width=0.13\textwidth]{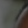} \\[-4pt] {\footnotesize (14,0)} \end{tabular} &
                \begin{tabular}{c} \includegraphics[width=0.13\textwidth]{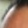} \\[-4pt] {\footnotesize (28,0)} \end{tabular} &
                \begin{tabular}{c} \includegraphics[width=0.13\textwidth]{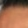} \\[-4pt] {\footnotesize (42,0)} \end{tabular} &
                \begin{tabular}{c} \includegraphics[width=0.13\textwidth]{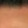} \\[-4pt] {\footnotesize (56,0)} \end{tabular} &
                \begin{tabular}{c} \includegraphics[width=0.13\textwidth]{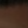} \\[-4pt] {\footnotesize (70,0)} \end{tabular} &
                \begin{tabular}{c} \includegraphics[width=0.13\textwidth]{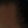} \\[-4pt] {\footnotesize (84,0)} \end{tabular} \\[7pt]
                
                \begin{tabular}{c} \includegraphics[width=0.13\textwidth]{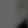} \\[-4pt] {\footnotesize (0,14)} \end{tabular} &
                \begin{tabular}{c} \includegraphics[width=0.13\textwidth]{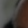} \\[-4pt] {\footnotesize (14,14)} \end{tabular} &
                \begin{tabular}{c} \includegraphics[width=0.13\textwidth]{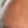} \\[-4pt] {\footnotesize (28,14)} \end{tabular} &
                \begin{tabular}{c} \includegraphics[width=0.13\textwidth]{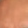} \\[-4pt] {\footnotesize (42,14)} \end{tabular} &
                \begin{tabular}{c} \includegraphics[width=0.13\textwidth]{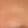} \\[-4pt] {\footnotesize (56,14)} \end{tabular} &
                \begin{tabular}{c} \includegraphics[width=0.13\textwidth]{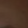} \\[-4pt] {\footnotesize (70,14)} \end{tabular} &
                \begin{tabular}{c} \includegraphics[width=0.13\textwidth]{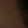} \\[-4pt] {\footnotesize (84,14)} \end{tabular} \\[7pt]
                
                \begin{tabular}{c} \includegraphics[width=0.13\textwidth]{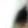} \\[-4pt] {\footnotesize (0,28)} \end{tabular} &
                \begin{tabular}{c} \includegraphics[width=0.13\textwidth]{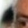} \\[-4pt] {\footnotesize (14,28)} \end{tabular} &
                \begin{tabular}{c} \includegraphics[width=0.13\textwidth]{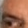} \\[-4pt] {\footnotesize (28,28)} \end{tabular} &
                \begin{tabular}{c} \includegraphics[width=0.13\textwidth]{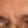} \\[-4pt] {\footnotesize (42,28)} \end{tabular} &
                \begin{tabular}{c} \includegraphics[width=0.13\textwidth]{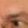} \\[-4pt] {\footnotesize (56,28)} \end{tabular} &
                \begin{tabular}{c} \includegraphics[width=0.13\textwidth]{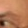} \\[-4pt] {\footnotesize (70,28)} \end{tabular} &
                \begin{tabular}{c} \includegraphics[width=0.13\textwidth]{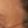} \\[-4pt] {\footnotesize (84,28)} \end{tabular} \\[7pt]
                
                \begin{tabular}{c} \includegraphics[width=0.13\textwidth]{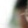} \\[-4pt] {\footnotesize (0,42)} \end{tabular} &
                \begin{tabular}{c} \includegraphics[width=0.13\textwidth]{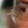} \\[-4pt] {\footnotesize (14,42)} \end{tabular} &
                \begin{tabular}{c} \includegraphics[width=0.13\textwidth]{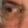} \\[-4pt] {\footnotesize (28,42)} \end{tabular} &
                \begin{tabular}{c} \includegraphics[width=0.13\textwidth]{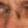} \\[-4pt] {\footnotesize (42,42)} \end{tabular} &
                \begin{tabular}{c} \includegraphics[width=0.13\textwidth]{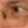} \\[-4pt] {\footnotesize (56,42)} \end{tabular} &
                \begin{tabular}{c} \includegraphics[width=0.13\textwidth]{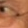} \\[-4pt] {\footnotesize (70,42)} \end{tabular} &
                \begin{tabular}{c} \includegraphics[width=0.13\textwidth]{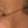} \\[-4pt] {\footnotesize (84,42)} \end{tabular} \\[7pt]
                
                \begin{tabular}{c} \includegraphics[width=0.13\textwidth]{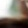} \\[-4pt] {\footnotesize (0,56)} \end{tabular} &
                \begin{tabular}{c} \includegraphics[width=0.13\textwidth]{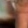} \\[-4pt] {\footnotesize (14,56)} \end{tabular} &
                \begin{tabular}{c} \includegraphics[width=0.13\textwidth]{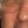} \\[-4pt] {\footnotesize (28,56)} \end{tabular} &
                \begin{tabular}{c} \includegraphics[width=0.13\textwidth]{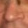} \\[-4pt] {\footnotesize (42,56)} \end{tabular} &
                \begin{tabular}{c} \includegraphics[width=0.13\textwidth]{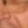} \\[-4pt] {\footnotesize (56,56)} \end{tabular} &
                \begin{tabular}{c} \includegraphics[width=0.13\textwidth]{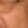} \\[-4pt] {\footnotesize (70,56)} \end{tabular} &
                \begin{tabular}{c} \includegraphics[width=0.13\textwidth]{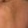} \\[-4pt] {\footnotesize (84,56)} \end{tabular} \\[7pt]
                
                \begin{tabular}{c} \includegraphics[width=0.13\textwidth]{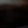} \\[-4pt] {\footnotesize (0,70)} \end{tabular} &
                \begin{tabular}{c} \includegraphics[width=0.13\textwidth]{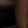} \\[-4pt] {\footnotesize (14,70)} \end{tabular} &
                \begin{tabular}{c} \includegraphics[width=0.13\textwidth]{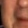} \\[-4pt] {\footnotesize (28,70)} \end{tabular} &
                \begin{tabular}{c} \includegraphics[width=0.13\textwidth]{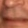} \\[-4pt] {\footnotesize (42,70)} \end{tabular} &
                \begin{tabular}{c} \includegraphics[width=0.13\textwidth]{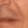} \\[-4pt] {\footnotesize (56,70)} \end{tabular} &
                \begin{tabular}{c} \includegraphics[width=0.13\textwidth]{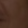} \\[-4pt] {\footnotesize (70,70)} \end{tabular} &
                \begin{tabular}{c} \includegraphics[width=0.13\textwidth]{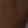} \\[-4pt] {\footnotesize (84,70)} \end{tabular} \\[7pt]

                \begin{tabular}{c} \includegraphics[width=0.13\textwidth]{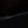} \\[-4pt] {\footnotesize (0,84)} \end{tabular} &
                \begin{tabular}{c} \includegraphics[width=0.13\textwidth]{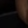} \\[-4pt] {\footnotesize (14,84)} \end{tabular} &
                \begin{tabular}{c} \includegraphics[width=0.13\textwidth]{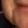} \\[-4pt] {\footnotesize (28,84)} \end{tabular} &
                \begin{tabular}{c} \includegraphics[width=0.13\textwidth]{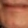} \\[-4pt] {\footnotesize (42,84)} \end{tabular} &
                \begin{tabular}{c} \includegraphics[width=0.13\textwidth]{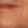} \\[-4pt] {\footnotesize (56,84)} \end{tabular} &
                \begin{tabular}{c} \includegraphics[width=0.13\textwidth]{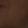} \\[-4pt] {\footnotesize (70,84)} \end{tabular} &
                \begin{tabular}{c} \includegraphics[width=0.13\textwidth]{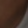} \\[-4pt] {\footnotesize (84,84)} \end{tabular} \\
            \end{tabular}
            
            \vspace{3pt}
            {\small (a) $28 \times 28$}
        \end{minipage} &
        \begin{minipage}{0.25\textwidth}
            \centering
            \setlength{\tabcolsep}{1.5pt}
            \begin{tabular}{ccc}
                \begin{tabular}{c} \includegraphics[width=0.32\textwidth]{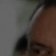} \\[-4pt] {\footnotesize (0,0)} \end{tabular} &
                \begin{tabular}{c} \includegraphics[width=0.32\textwidth]{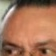} \\[-4pt] {\footnotesize (28,0)} \end{tabular} &
                \begin{tabular}{c} \includegraphics[width=0.32\textwidth]{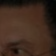} \\[-4pt] {\footnotesize (56,0)} \end{tabular} \\[7pt]
                
                \begin{tabular}{c} \includegraphics[width=0.32\textwidth]{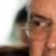} \\[-4pt] {\footnotesize (0,28)} \end{tabular} &
                \begin{tabular}{c} \includegraphics[width=0.32\textwidth]{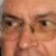} \\[-4pt] {\footnotesize (28,28)} \end{tabular} &
                \begin{tabular}{c} \includegraphics[width=0.32\textwidth]{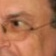} \\[-4pt] {\footnotesize (56,28)} \end{tabular} \\[7pt]
                
                \begin{tabular}{c} \includegraphics[width=0.32\textwidth]{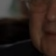} \\[-4pt] {\footnotesize (0,56)} \end{tabular} &
                \begin{tabular}{c} \includegraphics[width=0.32\textwidth]{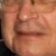} \\[-4pt] {\footnotesize (28,56)} \end{tabular} &
                \begin{tabular}{c} \includegraphics[width=0.32\textwidth]{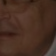} \\[-4pt] {\footnotesize (56,56)} \end{tabular} \\
            \end{tabular}
            
            \vspace{3pt}
            {\small (b) $56 \times 56$}
        \end{minipage} &
        \begin{minipage}{0.2\textwidth}
            \centering
            \includegraphics[width=0.95\textwidth]{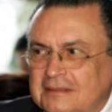}
            
            \vspace{3pt}
            {\small (c) $112 \times 112$}
        \end{minipage}
    \end{tabular}
    \end{adjustbox}
    
    \caption{Restricted receptive fields of sizes (a) $28 \times 28$ and (b) $56 \times 56$ are shown for a given (c) $112 \times 112$ face image. The top-left coordinates of each image patch are indicated below the corresponding patch. For RRFNet-28, the four patches at the corners are excluded, while for RRFNet-56, one patch at each corner is excluded.}
    \label{fig:receptive_fields}
\end{figure*}

%% file: figures/heatmap.tex
\begin{figure*}[t]
\centering
\setlength{\tabcolsep}{6pt}
\renewcommand{\arraystretch}{1.0}
\begin{tabular}{ccccc|ccccc}
\textbf{Image A} & \textbf{Image B} & \textbf{Patch A} & \textbf{Patch B} & \textbf{Score} &
\textbf{Image A} & \textbf{Image B} & \textbf{Patch A} & \textbf{Patch B} & \textbf{Score} \\
\hline
\includegraphics[width=0.08\linewidth]{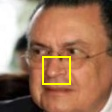} &
\includegraphics[width=0.08\linewidth]{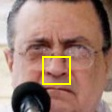} &
\raisebox{.5\height}{\includegraphics[width=0.04\linewidth]{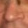}} &
\raisebox{.5\height}{\includegraphics[width=0.04\linewidth]{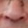}} &
\raisebox{2.2\height}{\makebox[0pt]{\textbf{22.38}}} &
\includegraphics[width=0.08\linewidth]{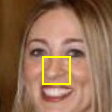} &
\includegraphics[width=0.08\linewidth]{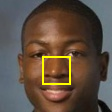} &
\raisebox{.5\height}{\includegraphics[width=0.04\linewidth]{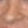}} &
\raisebox{.5\height}{\includegraphics[width=0.04\linewidth]{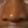}} &
\raisebox{2.2\height}{\makebox[0pt]{\textbf{-9.82}}} \\[2pt]
\includegraphics[width=0.08\linewidth]{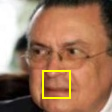} &
\includegraphics[width=0.08\linewidth]{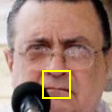} &
\raisebox{.5\height}{\includegraphics[width=0.04\linewidth]{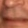}} &
\raisebox{.5\height}{\includegraphics[width=0.04\linewidth]{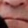}} &
\raisebox{2.2\height}{\makebox[0pt]{\textbf{20.64}}} &
\includegraphics[width=0.08\linewidth]{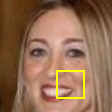} &
\includegraphics[width=0.08\linewidth]{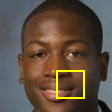} &
\raisebox{.5\height}{\includegraphics[width=0.04\linewidth]{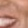}} &
\raisebox{.5\height}{\includegraphics[width=0.04\linewidth]{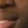}} &
\raisebox{2.2\height}{\makebox[0pt]{\textbf{-3.14}}} \\[2pt]
\includegraphics[width=0.08\linewidth]{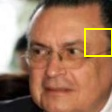} &
\includegraphics[width=0.08\linewidth]{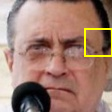} &
\raisebox{.5\height}{\includegraphics[width=0.04\linewidth]{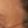}} &
\raisebox{.5\height}{\includegraphics[width=0.04\linewidth]{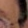}} &
\raisebox{2.2\height}{\makebox[0pt]{\textbf{9.63}}} &
\includegraphics[width=0.08\linewidth]{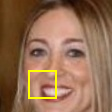} &
\includegraphics[width=0.08\linewidth]{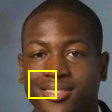} &
\raisebox{.5\height}{\includegraphics[width=0.04\linewidth]{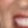}} &
\raisebox{.5\height}{\includegraphics[width=0.04\linewidth]{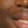}} &
\raisebox{2.2\height}{\makebox[0pt]{\textbf{-1.93}}} \\[2pt]
\includegraphics[width=0.08\linewidth]{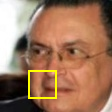} &
\includegraphics[width=0.08\linewidth]{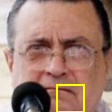} &
\raisebox{.5\height}
{\includegraphics[width=0.04\linewidth]{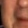}} &
\raisebox{.5\height}
{\includegraphics[width=0.04\linewidth]{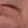}} &
\raisebox{2.2\height}{\makebox[0pt]{\textbf{4.45}}} &
\includegraphics[width=0.08\linewidth]{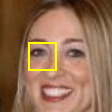} &
\includegraphics[width=0.08\linewidth]{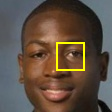} &
\raisebox{.5\height}{\includegraphics[width=0.04\linewidth]{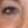}} &
\raisebox{.5\height}{\includegraphics[width=0.04\linewidth]{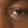}} &
\raisebox{2.2\height}{\makebox[0pt]{\textbf{-1.51}}} \\[2pt]
\includegraphics[width=0.08\linewidth]{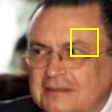} &
\includegraphics[width=0.08\linewidth]{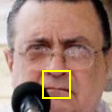} &
\raisebox{.5\height}
{\includegraphics[width=0.04\linewidth]{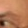}} &
\raisebox{.5\height}
{\includegraphics[width=0.04\linewidth]{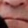}} &
\raisebox{2.2\height}{\makebox[0pt]{\textbf{-1.04}}} &
\includegraphics[width=0.08\linewidth]{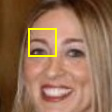} &
\includegraphics[width=0.08\linewidth]{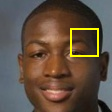} &
\raisebox{.5\height}{\includegraphics[width=0.04\linewidth]{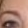}} &
\raisebox{.5\height}{\includegraphics[width=0.04\linewidth]{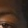}} &
\raisebox{2.2\height}{\makebox[0pt]{\textbf{4.03}}} \\[2pt]
\includegraphics[width=0.08\linewidth]{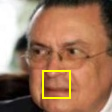} &
\includegraphics[width=0.08\linewidth]{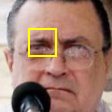} &
\raisebox{.5\height}
{\includegraphics[width=0.04\linewidth]{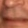}} &
\raisebox{.5\height}
{\includegraphics[width=0.04\linewidth]{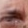}} &
\raisebox{2.2\height}{\makebox[0pt]{\textbf{-1.07}}} &

\includegraphics[width=0.08\linewidth]{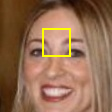} &
\includegraphics[width=0.08\linewidth]{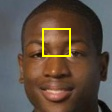} &
\raisebox{.5\height}{\includegraphics[width=0.04\linewidth]{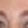}} &
\raisebox{.5\height}{\includegraphics[width=0.04\linewidth]{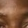}} &
\raisebox{2.2\height}{\makebox[0pt]{\textbf{4.18}}} \\[2pt]
\vdots & \vdots & \vdots & \vdots & \vdots &
\vdots & \vdots & \vdots & \vdots & \vdots \\[2pt]
 &  &  &  & \hfill$+$ &
 &  &  &  & \hfill$+$ \\
\cline{5-5} \cline{10-10}
\multicolumn{4}{c}{\textbf{Face Similarity Score}} & \textbf{0.70} &
\multicolumn{4}{c}{\textbf{Face Similarity Score}} & \textbf{0.02} \\[6pt]
\multicolumn{5}{c}{%
\begin{tabular}{@{}c@{\hspace{3pt}}c@{}}
\raisebox{0\height}{\includegraphics[height=0.15\linewidth]{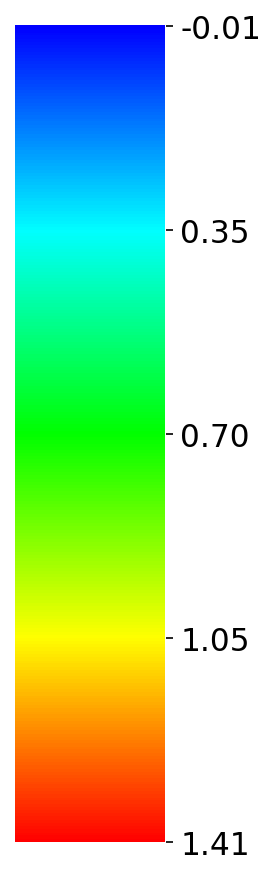}} &
\includegraphics[width=0.30\linewidth]{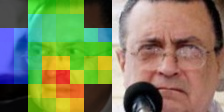}
\end{tabular}
} &
\multicolumn{5}{c}{%
\begin{tabular}{@{}c@{\hspace{3pt}}c@{}}
\raisebox{0\height}{\includegraphics[height=0.15\linewidth]{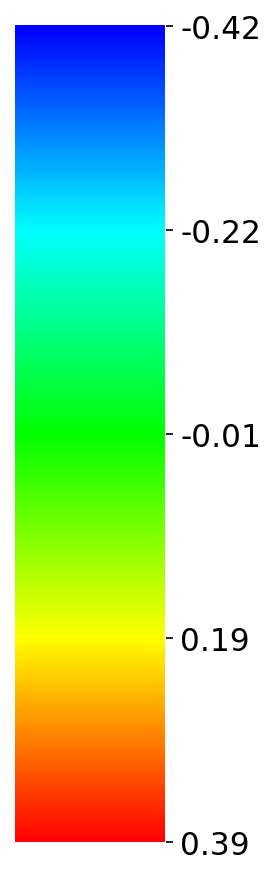}} &
\includegraphics[width=0.30\linewidth]{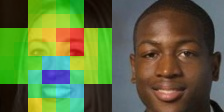}
\end{tabular}
} \\[2pt]
\end{tabular}
\caption{Visualization of patch-level similarities for two image pairs computed using RRFNet-28. Each row presents a pair of corresponding patches (Patch A and Patch B) from Image A and Image B, along with their similarity scores. The overall face similarity score is obtained by aggregating the scores from all patch pairs. The heatmaps at the bottom illustrate the spatial distribution of patch similarities on Image A.}
\label{fig:patch_similarity}
\end{figure*}

%% file: sec/3_method.tex
\section{Methodology}
\label{sec:method}

In this work, we propose a distance metric to decompose face similarity measure into local regions. For an image size of $W \times H$, we choose $w<W$ and $h<H$ to learn local representations for each $w \times h$ restricted patch. The image patches are uniformly distributed across the image. Two approaches are considered for learning local representations and constructing global image similarity metric: (i) training region-based CNNs for each $k$ patches (see Section \ref{sec:region-based}) and (ii) single CNN to learn a global image representation as the mean of local representations (see Section \ref{sec:rrfnet}). In the first approach, we adopt the same training objective as in \cite{deng2019arcface} to learn local features for each patch obtained from $W \times H$ face images. After the CNNs are trained, a second training phase learns weights for each region similarity using logistic regression. Weighted sum of $k$ local similarity scores, obtained from corresponding regions between two images, is used to the build global distance metric. In the second approach, rather than training separate CNNs, a single CNN is trained and the global representation is computed as the mean of the patch-level features, enabling face similarity to be measured directly from patch representations. We use the terms \textbf{patch} and \textbf{restricted receptive field} interchangeably throughout the manuscript to describe our approach.

\subsection{Patch Representation Learning}
\label{sec:region-based}

We first divide each face image into a set of smaller, uniformly distributed patches. Each patch is treated as an independent unit, allowing the model to capture fine-grained localized feature representations. Unlike the traditional approach of generating a single \( N \)-dimensional representation for an entire image, we obtain \( (k, N) \)-dimensional representation where a face image is divided into $k$ patches. Patches are defined with position and window size. For a given face image, we extract \( k \) patches, which we will refer to as \( P_1, P_2, \dots, P_k \), each corresponding to a distinct region and size of the image. While $112 \times 112$ is traditionally used image size for face recognition research in recent works, we analyze two patch sizes, $28 \times 28$ and $56 \times 56$ in our experiments, demonstrated in Fig. \ref{fig:receptive_fields}. 

To ensure a fair comparison, we adopt the same CNN architecture and loss function as in \cite{deng2019arcface}. The only modification is changing the stride in the first ResNet block from $2$ to $1$, preventing the receptive field in each ResNet \cite{he2016deep} block from shrinking aggressively when using smaller input sizes ($28 \times 28$ and $56 \times 56$) for patch representation learning. For each of the $k$ extracted patches of size $w \times h$, we train a separate CNN.

\input{algos/RRFNet}

\textbf{Region-based Similarity Metric.}
After local feature learning phase is completed, we train a logistic regression model to learn the weights \( w_1, w_2, \dots, w_k \) for each patch similarity between two images. The weights are learned on the same training set used in the first phase to distinguish genuine and impostor pairs. These weights allow the model to assign importance to different parts, thereby enhancing the model's capacity to focus on the most discriminative regions of the image to make a binary decision.

Let the output of each CNN, trained in the first step, corresponding to the \( i \)-th patch be \( f_i \), which is a \( N \)-dimensional feature vector representing patch \( P_i \). Local similarity between two images is computed for each corresponding patches \( P_i^A \) from image \( A \) and \( P_i^B \) image \( B \) using cosine similarity:
\begin{equation}
S_{local}(P_i^A, P_i^B) = \frac{f_i^A \cdot f_i^B}{\|f_i^A\| \|f_i^B\|}
\end{equation}
where \( f_i^A \) and \( f_i^B \) are the feature vectors of the \( i \)-th patch from images \( A \) and \( B \), respectively. 

The global similarity metric between the two images is then defined as a weighted sum of the local similarity scores:
\begin{equation}
S_{global}(A, B) = \sum_{i=1}^k w_i \cdot S_{local}(P_i^A, P_i^B)
\end{equation}

\subsection{Restricted Receptive Field Network (RRFNet)}
\label{sec:rrfnet}

In this approach, we demonstrate that by making only minor modifications to a standard CNN backbone, it is possible to learn a compact feature space shared across all restricted receptive fields, enabling verification through patch-level comparisons. Following the training setup of \cite{deng2019arcface}, instead of processing the entire $112 \times 112$ face image to obtain a global representation, we extract features from $28 \times 28$ (RRFNet-28) and $56 \times 56$ (RRFNet-56) image patches. While this approach constrains the receptive field of the network, it enhances the transparency of verification decisions by decomposing similarity into local regions.

Comparison of the proposed RRFNet approach with the ResNet \cite{he2016deep} is depicted in Table \ref{rrfnet_vs_resnet}. While traditional approach with ResNet produce a $512$-dimensional representation from the entire face image, our approach extracts $512$-dimensional representations for each restricted receptive field. At the end of the RRFNet, a global representation is obtained as the mean of the patch representations enabling expression of the global similarity as patch similarities (illustrated in Fig. \ref{fig:patch_similarity}).

\textbf{Patch-level Similarity Metric.} 
Typically, the cosine similarity between two $512$-dimensional face representations, extracted from a pretrained recognition model, is used for face verification. In our approach, the global face representation is defined as the mean of the patch-level representations. This formulation implies that the similarity between two images can equivalently be expressed in terms of the sum of dot products between all pairs of their patch representations.

Let $\{\mathbf{f}_{i}^{A} \in \mathbb{R}^{512}\}_{i=1}^{K}$ and $\{\mathbf{f}_{i}^{B} \in \mathbb{R}^{512}\}_{i=1}^{K}$ denote the patch-level feature vectors extracted from two face images, where $K$ is the total number of patches. Here, $\mathbf{f}_{i}^{A}$ and $\mathbf{f}_{i}^{B}$ represent the feature of the $i$-th patch in images $A$ and $B$, respectively.

The global representation of each image is obtained by averaging its patch features:
\begin{equation}\label{eq:global_mean}
\mathbf{F}^{A} = \frac{1}{K} \sum_{i=1}^{K} \mathbf{f}_{i}^{A},
\qquad
\mathbf{F}^{B} = \frac{1}{K} \sum_{i=1}^{K} \mathbf{f}_{i}^{B}.
\end{equation}

The cosine similarity between the two global representations is defined as
\begin{equation}\label{eq:cos_global}
\operatorname{sim}(\mathbf{F}^{A}, \mathbf{F}^{B})
= \frac{ \mathbf{F}^{A} \cdot \mathbf{F}^{B} }
{ \|\mathbf{F}^{A}\| \, \|\mathbf{F}^{B}\| }.
\end{equation}

Substituting \eqref{eq:global_mean} into the numerator of \eqref{eq:cos_global}, we obtain
\begin{align}\label{eq:num_expanded}
\mathbf{F}^{A} \cdot \mathbf{F}^{B}
&= \left( \frac{1}{K} \sum_{i=1}^{K} \mathbf{f}_{i}^{A} \right)
   \cdot
   \left( \frac{1}{K} \sum_{j=1}^{K} \mathbf{f}_{j}^{B} \right) \nonumber \\
&= \frac{1}{K^{2}}
   \sum_{i=1}^{K} \sum_{j=1}^{K}
   \mathbf{f}_{i}^{A} \cdot \mathbf{f}_{j}^{B}.
\end{align}

And the norms of the global representations can be written as
\begin{align}
\|\mathbf{F}^{A}\|
&= \frac{1}{K}
\sqrt{
\sum_{i=1}^{K} \sum_{j=1}^{K}
\mathbf{f}_{i}^{A} \cdot \mathbf{f}_{j}^{A}
}, \label{eq:normA_sqrt} \\[4pt]
\|\mathbf{F}^{B}\|
&= \frac{1}{K}
\sqrt{
\sum_{i=1}^{K} \sum_{j=1}^{K}
\mathbf{f}_{i}^{B} \cdot \mathbf{f}_{j}^{B}
}. \label{eq:normB_sqrt}
\end{align}

Substituting \eqref{eq:num_expanded}--\eqref{eq:normB_sqrt} into \eqref{eq:cos_global}, the scalar factors of $\tfrac{1}{K^{2}}$ cancel out, yielding a similarity formulation expressed entirely in terms of patch-level dot products:
\begin{equation}\label{eq:cos_allpairs}
\operatorname{sim}(\mathbf{F}^{A}, \mathbf{F}^{B})
=
\frac{
\displaystyle
\sum_{i=1}^{K} \sum_{j=1}^{K}
\mathbf{f}_{i}^{A} \cdot \mathbf{f}_{j}^{B}
}{
\sqrt{
\displaystyle
\sum_{i=1}^{K} \sum_{j=1}^{K}
\mathbf{f}_{i}^{A} \cdot \mathbf{f}_{j}^{A}
}
\;
\sqrt{
\displaystyle
\sum_{i=1}^{K} \sum_{j=1}^{K}
\mathbf{f}_{i}^{B} \cdot \mathbf{f}_{j}^{B}
}
}.
\end{equation}

%% file: algos/RRFNet.tex
\begin{table*}[t]
\caption{Comparison of RRFNet and ResNet architectures.}
\centering
\renewcommand{\arraystretch}{1.3}
\setlength{\tabcolsep}{12pt}
\begin{tabular}{>{\raggedright\arraybackslash}m{6cm} >{\raggedright\arraybackslash}m{6cm}}
\toprule
\textbf{RRFNet} & \textbf{ResNet} \\
\midrule
\multicolumn{2}{l}{\textit{$B$: Batch size,\ $W$: Image width,\ $H$: Image height,\ $w$: Patch width,\ $h$: Patch height,\ $k$: Number of patches}} \\
\\
\textbf{Input image:} $B,W,H,C$ (e.g., $1,112,112,3$) & \textbf{Input image:} $B,W,H,C$ (e.g., $1,112,112,3$) \\
\textbf{Create patches:} $k \times B,w,h,C$ (e.g., $33,28,28,3$) & \\
\textbf{Block1:} $k \times B,w,h,64$ & \textbf{Block1:} $B,W/2,H/2,64$ \\
\textbf{Block2:} $k \times B,w/2,h/2,128$ & \textbf{Block2:} $B,W/4,H/4,128$ \\
\textbf{Block3:} $k \times B,w/4,h/4,256$ & \textbf{Block3:} $B,W/8,H/8,256$ \\
\textbf{Block4:} $k \times B,w/8,h/8,512$ (e.g., $33,4,4,512$)& \textbf{Block4:} $B,W/16,H/16,512$ (e.g., $1,7,7,512$)\\
\textbf{FC:} $k \times B,512$ (e.g., $33,512$)& \textbf{FC:} $B,512$ (e.g., $1,512$)\\
\textbf{Mean:} $B,512$ & \\
\bottomrule
\end{tabular}
\label{rrfnet_vs_resnet}
\end{table*}

%% file: sec/4_results.tex
\section{Experiment}
\label{experiment}

\textbf{Dataset.}
We evaluate the verification performance of the proposed method on seven benchmark datasets. LFW \cite{huang2008labeled}, CFP-FP \cite{sengupta2016frontal}, CPLFW \cite{zheng2018cross}, CALFW \cite{zheng2017cross}, and AGEDB \cite{moschoglou2017agedb} are commonly used in recent face recognition research for assessing robustness to age and pose variations. In addition, the Eclipse (ECL) and Hadrian (HAD) datasets \cite{wu2024goldilocks} are used to evaluate performance under variations in illumination and facial hair.

In our experiments, we use two receptive field sizes, $56 \times 56$ and $28 \times 28$, on $112 \times 112$ face images to evaluate the verification performance. We first analyze the verification performance of individual patches by training the CNNs described in Section \ref{sec:region-based}. Each model is trained for 20 epochs on the WebFace4M dataset \cite{zhu2021webface260m}, using the ResNet100 architecture \cite{he2016deep}. For the $56 \times 56$ patches, the network retains the same number of parameters as in \cite{deng2019arcface} with $112 \times 112$ inputs, as we use a stride of 1 instead of 2 in the first ResNet block. On the other hand, for the $28 \times 28$ patch size, the feature map reduces to $4 \times 4$ in the last ResNet block, resulting in fewer parameters in the fully-connected layer compared to the baseline approach \cite{deng2019arcface}.

\subsection{Verification Rates for Region-based Similarity.}
Individual verification performance of the each receptive field is given in Table~\ref{tab:patch_ver_val}. Note that, while there are forty-nine different positions for $28 \times 28$ patches, shown in Fig. \ref{fig:receptive_fields}, twenty-eight CNNs are trained because symmetric patches on the left and right sides share the same network due to facial symmetry. Similarly six networks are used for $56 \times 56$ patches instead of nine. For example, in Fig. \ref{fig:receptive_fields}, patches at $(28,28)$ and $(56,28)$ trained with the same CNNs. As we apply horizontal flipping to augment representations during inference following \cite{insightface_github}, only left-side positions are depicted in the Table~\ref{tab:patch_ver_val}.

While patches in the middle region of the face achieve the highest accuracies, the lower half of the face outperforms the upper half across most datasets. However, on Hadrian, patches in the upper regions perform better than those in the lower regions due to significant beard and mustache variations between image pairs. Although smaller patch sizes lead to an expected drop in accuracy, surprisingly, a single $56 \times 56$ patch at position $(28,28)$ achieves verification rates comparable to the full $112 \times 112$ receptive field.

\textbf{Combination of Receptive Fields.} 
Score-level combination of the receptive fields is used for verification decisions. To compare the performance of the different receptive field sizes, combinations are performed in 3 ways: only $28 \times 28$ patches, only $56 \times 56$ and combination of $28 \times 28$ and $56 \times 56$. Verification rates using the full image size $112 \times 112$ \cite{deng2019arcface}, are also given for comparison. Results are shown at the bottom of Tab \ref{tab:patch_ver_val}.

\input{tables/val_verification}

\input{tables/results_new}

\input{tables/cross_pose}

As the input size decreases from $112 \times 112$ to $28 \times 28$, the score-level combination of region-based similarities using $28 \times 28$ patches yields lower accuracies, as expected. In contrast, $56 \times 56$ patches achieve competitive results, and combining both patch sizes further improves accuracy. While using the entire receptive field ($112 \times 112$) is more effective on datasets with pose variation (CFPFP and CPLFW), our region-based similarity approach improves performance on frontal image pairs (ECL and HAD) significantly.

\subsection{Verification Rates for RRFNet}
While competitive performance is achieved with $56 \times 56$ patches, a notable performance drop for cross-pose datasets is observed for $28 \times 28$ patches in Table~\ref{tab:patch_ver_val} when using region-based similarities. We hypothesize that this primarily arises from restricting comparisons to corresponding patches at the same spatial positions. To overcome this limitation, we introduce a second approach, RRFNet, which compares each patch in one image with all patches in the other image to compute a comprehensive similarity score. For instance, RRFNet-28 divides each $112 \times 112$ face image into 33 patches of size $28 \times 28$, resulting in $33 \times 33 = 1089$ patch pairs used to measure image-level similarity.

To determine the receptive fields for RRFNet-56, we evaluate four different configurations for the $56 \times 56$ setting, as shown in Table~\ref{tab:patchconf}. The highest verification rates are obtained using 5 patches located at $(0,28)$, $(28,0)$, $(56,28)$, $(28,56)$, and $(28,28)$ in our preliminarily experiments. To balance accuracy and computational cost, we do not consider configurations with more patches, and this 5-patch setup is used to report results. Note, this choice excludes the four $28 \times 28$ corner regions of the $112 \times 112$ face images as shown in Fig. \ref{fig:receptive_fields}. For RRFNet-28, which uses $28 \times 28$ patches, the same positions are applied, yielding a total of 33 patches.

Verification performance of the proposed approach is given in Table~\ref{tab:results_new}. Publicly-available models of three recent approaches \cite{kim2022adaface, uniface, vitkprpe} are used for a comparison. As we use the same architecture and the loss function, we trained models for the ArcFace \cite{deng2019arcface} in the same settings to report results for a fair comparison. Our RRFNet-28 and RRFNet-56 models trained for 30 epochs. During training $[-5,5]$ vertical and horizontal shifts are applied as data augmentation. Also, effect of masked patch augmentation is analyzed with two masked patch ratio $20\%$ and $40\%$. Two datasets, Casia-WebFace \cite{yi2014learning} and WebFace4M \cite{zhu2021webface260m}, and two CNN architectures, ResNet50 and ResNet100 \cite{he2016deep}, are used for training recognition models using the publicly-available implementation \cite{insightface_github}.

\input{tables/patch56selection}

\input{tables/inference_time}

\input{figures/del_imp_gen_fig}
\input{figures/fpr_and_fnr}

As shown in Table \ref{tab:results_new}, RRFNet-56 consistently achieves the highest verification rates across 7 datasets under different training settings. The only exception is CPLFW, where KP-RPE \cite{vitkprpe} obtains $95.40$ accuracy. While KP-RPE requires an additional model for keypoint supervision, our architecture does not rely on any positional information and processes all patches uniformly. Surprisingly, we observe that RRFNet-28 achieves competitive results, even though verification is performed using patches as small as $28 \times 28$. Note, the verification rates achieved with $28 \times 28$ patches show a significant improvement over the region-based similarity approach (see Table~\ref{tab:patch_ver_val} comb. $28 \times 28$). To further investigate this gain, we measure the percentage of patches whose most similar match in genuine pairs are located at the same spatial position. Results are reported in Table~\ref{tab:cross_pose_top_match} for RRFNet-28. We observe that this ratio can drop to as low as $20\%$ in cross-pose scenarios, highlighting the importance of cross-region patch matching compared to the region-based similarity approach.

%% file: tables/val_verification.tex
\setlength{\tabcolsep}{5.2pt} 
\renewcommand{\arraystretch}{1.3}
\begin{table*}[hbt!]
\caption{Face verification accuracy (\%) across seven benchmark datasets for each receptive field size and position individually. Additionally, score-level combination of receptive fields are given at the bottom. Results are reported for receptive fields of sizes $28 \times 28$, $56 \times 56$, and $112 \times 112$, with coordinates $(x,y)$ indicating the top-left corner position. The best-performing receptive field for $56 \times 56$ and the top-3 performing receptive fields for $28 \times 28$ are highlighted in \textbf{bold} for each dataset.} 
\centering
\begin{tabular}{|c|c||c|c|c|c|c|c|c||c|}
\hline
\multicolumn{2}{|c||}{Receptive Fields} & \multicolumn{8}{c|}{Dataset} \\
\hline
Size & Position & LFW & CFPFP & CPLFW & AGEDB & CALFW & ECL & HAD & AVG \\
\hline
\multirow{28}{*}{$28 \times 28$}&(0,0) & $82.45$ & $56.31$ & $63.07$ & $65.30$ & $56.43$ & $50.02$ & $55.83$ & $61.34$ \\
&(14,0) & $90.50$ & $63.07$ & $68.93$ & $72.87$ & $62.45$ & $50.28$ & $63.88$ & $67.43$ \\
&(28,0)& $90.13$ & $64.44$ & $71.73$ & $73.42$ & $63.47$ & $49.37$ & $61.47$ & $67.72$ \\
& (42,0) & $89.43$ & $64.76$ & $71.78$ & $72.97$ & $63.65$ & $53.85$ & $64.72$ & $68.74$ \\
\cline{2-10}
& (0,14) & $87.32$ & $68.74$ & $65.97$ & $70.40$ & $61.18$ & $50.53$ & $60.90$ & $66.43$ \\
& (14,14) & $94.73$ & $77.20$ & $71.65$ & $81.10$ & $74.53$ & $58.35$ & $73.82$ & $75.91$ \\
& (28,14) & $94.73$ & $80.44$ & $77.55$ & $83.12$ & $77.77$ & $58.27$ & $74.27$ & $78.02$ \\
& (42,14) & $93.68$ & $79.84$ & $77.98$ & $80.73$ & $76.35$ & $60.77$ & $78.38$ & $78.25$ \\
\cline{2-10}
& (0,28) & $92.98$ & $76.01$ & $70.17$ & $78.77$ & $68.58$ & $55.42$ & $67.37$ & $72.76$ \\
& (14,28) & $98.22$ & $84.23$ & $76.85$ & $90.47$ & $84.17$ & $66.45$ & $80.58$ & $83.00$ \\
& (28,28) & $98.32$ & $88.84$ & $85.00$ & \bm{$92.43$} & $90.05$ & $69.67$ & \bm{$86.48$} & $87.26$ \\
& (42,28) & $97.77$ & $89.31$ & $83.25$ & $90.93$ & $88.93$ & $70.87$ & $84.68$ & $86.53$ \\
\cline{2-10}
& (0,42) & $94.32$ & $80.73$ & $71.27$ & $80.88$ & $69.62$ & $59.77$ & $70.27$ & $75.27$ \\
& (14,42) & $97.73$ & $82.53$ & $77.33$ & $90.63$ & $84.83$ & $67.68$ & $79.68$ & $82.92$ \\
& (28,42) & \bm{$98.60$} & $89.67$ & \bm{$85.13$} & \bm{$93.22$} & \bm{$91.33$} & $72.95$ & $84.90$ & \bm{$87.97$} \\
& (42,42) & \bm{$98.53$} & $90.17$ & \bm{$85.53$} & $92.42$ & \bm{$90.77$} & $73.37$ & \bm{$88.98$} & \bm{$88.25$} \\
\cline{2-10}
& (0,56) & $89.40$ & $72.56$ & $64.03$ & $75.37$ & $63.88$ & $55.82$ & $62.58$ & $69.38$ \\
& (14,56) & $96.30$ & $80.50$ & $73.28$ & $85.08$ & $77.47$ & $62.05$ & $69.00$ & $77.67$ \\
& (28,56) & \bm{$98.80$} & \bm{$92.71$} & $84.98$ & \bm{$92.52$} & $89.58$ & $70.72$ & $79.30$ & $86.23$ \\
& (42,56) & $98.40$ & \bm{$92.50$} & \bm{$86.75$} & $91.73$ & \bm{$91.58$} & \bm{$76.03$} & \bm{$86.60$} & \bm{$89.08$} \\
\cline{2-10}
& (0,70) & $85.13$ & $69.41$ & $61.88$ & $70.83$ & $60.77$ & $49.28$ & $51.77$ & $64.15$ \\
& (14,70) & $96.23$ & $80.80$ & $71.13$ & $85.38$ & $76.48$ & $63.20$ & $58.55$ & $75.97$ \\
& (28,70) & $98.45$ & $91.83$ & $84.10$ & $92.17$ & $89.08$ & \bm{$73.78$} & $70.93$ & $85.76$ \\
& (42,70) & $98.15$ & \bm{$92.31$} & $84.77$ & $91.67$ & $90.53$ & \bm{$76.83$} & $79.53$ & $87.68$ \\
\cline{2-10}
& (0,84) & $75.43$ & $63.56$ & $58.38$ & $61.05$ & $55.28$ & $49.95$ & $49.98$ & $59.09$ \\
& (14, 84) & $94.65$ & $79.57$ & $69.75$ & $81.73$ & $71.80$ & $56.30$ & $50.13$ & $71.99$ \\
& (28, 84) & $97.12$ & $89.17$ & $80.90$ & $89.42$ & $84.53$ & $69.88$ & $65.05$ & $82.30$ \\
& (42, 84) & $95.52$ & $87.89$ & $79.72$ & $87.55$ & $84.17$ & $73.02$ & $69.60$ & $82.50$ \\
\hline
\multirow{6}{*}{$56 \times 56$}& (0,0) & $99.15$ & $94.83$ & $90.22$ & $92.52$ & $93.93$ & $72.88$ & $89.50$ & $90.15$ \\
& (28,0) & $99.42$ & $95.77$ & $90.22$ & $94.98$ & $94.45$ & $77.70$ & $93.72$ & $92.32$ \\
\cline{2-10}
& (0,28) & $99.67$ & $97.91$ & \bm{$92.62$} & $96.32$ & $95.70$ & $78.75$ & $90.20$ & $93.02$ \\
& (28,28) & \bm{$99.72$} & \bm{$98.79$} & $92.48$ & \bm{$97.48$} & \bm{$95.90$} & $82.02$ & \bm{$94.45$} & \bm{$94.41$} \\
\cline{2-10}
& (0,56) & $99.63$ & $97.76$ & $90.87$ & $95.55$ & $94.92$ & $81.08$ & $80.98$ & $91.54$ \\
& (28,56) & $99.65$ & $98.77$ & $91.87$ & $97.03$ & $95.23$ & \bm{$82.22$} & $88.17$ & $93.28$ \\
\hline
$112 \times 112$ & - & \bm{$99.82$} & \bm{$99.27$} & \bm{$94.50$} & $98.02$ & $95.98$ & $84.35$ & $93.70$ & $95.09$ \\
\hline
\hline
\makecell{Comb. \\ $28 \times 28$} & - & $99.67$ & $96.99$ & $90.32$ & $96.03$ & $95.20$ & $80.05$ & $93.60$ & $93.12$ \\
\hline
\makecell{Comb. \\ $56 \times 56$} & - & $99.82$ & $99.10$ & $93.70$ & $97.82$ & $95.97$ & $84.20$ & $95.65$ & $95.18$ \\
\hline
\makecell{Comb. \\ $28 \times 28$ \\ \& \\ $56 \times 56$} & - & $99.73$ & $99.17$ & $93.65$ & \bm{$98.03$} & \bm{$96.03$} & \bm{$85.12$} & \bm{$96.87$} & \bm{$95.51$} \\
\hline

\end{tabular}
\label{tab:patch_ver_val}
\end{table*}

%% file: tables/results_new.tex
\setlength{\tabcolsep}{4pt} 
\renewcommand{\arraystretch}{1.2}

\begin{table*}[hbt!]
\caption{Comparison of state-of-the-art methods with the proposed patch-level similarity approach. Verification rates are reported on seven datasets. Results for RRFNet-56 under two training settings (training on CASIA-WebFace with ResNet50 and on WebFace4M with ResNet100) show improved verification rates over the state-of-the-art methods.} 
\centering
\begin{tabular}{|c|c|c||c|c|c|c|c|c|c||c|}
\hline
\multirow{2}{*}{Method} & \multirow{2}{*}{Backbone} & \multirow{2}{*}{Train Data} & \multicolumn{8}{c|}{Dataset} \\
\cline{4-11}
 & & & LFW & CFPFP & CPLFW & AGEDB & CALFW & ECL & HAD & AVG\\
\hline
UniFace \cite{uniface} & ResNet50 & Casia-WebFace & $99.57$ & $97.04$ & $90.58$ & $95.27$ & $94.33$ & $73.80$ & $82.13$ & $90.39$ \\
AdaFace \cite{kim2022adaface} & ResNet50 & Casia-WebFace & $99.42$ & $96.44$ & $90.02$ & $94.38$ & $93.43$ & $73.18$ & $80.25$ & $89.59$ \\
ArcFace \cite{deng2019arcface} & ResNet50 & Casia-WebFace & $99.37$ & $97.24$ & $90.33$ & $94.93$ & $93.47$ & $72.57$ & $81.23$ & $89.88$ \\
BagNet-17 \cite{brendel2018approximating} & ResNet50 & Casia-WebFace & $98.35$ & $92.53$ & $83.57$ & $89.32$ & $90.18$ & $60.87$ & $67.82$ & $83.23$ \\
BagNet-33 \cite{brendel2018approximating} & ResNet50 & Casia-WebFace & $98.98$ & $94.93$ & $87.33$ & $92.63$ & $92.68$ & $66.10$ & $72.27$ & $86.42$ \\
\makecell{\textbf{RRFNet-28}\\(40\% mask)} & ResNet50 & Casia-WebFace & $99.33$ & $97.71$ & $90.37$ & $95.18$ & $93.77$ & $72.28$ & $81.73$ & $90.05$ \\
\makecell{\textbf{RRFNet-28}\\(20\% mask)} & ResNet50 & Casia-WebFace & $99.43$ & $97.61$ & $90.13$ & $94.73$ & $94.03$ & $73.57$ & $83.43$ & $90.42$ \\
\makecell{\textbf{RRFNet-28}\\(w/o mask)} & ResNet50 & Casia-WebFace & $99.33$ & $97.73$ & $89.73$ & $95.28$ & $94.22$ & $73.40$ & $81.00$ & $90.10$ \\
\makecell{\textbf{RRFNet-56}\\(40\% mask)} & ResNet50 & Casia-WebFace & $99.48$ & \bm{$98.11$} & $90.28$ & $95.70$ & $94.25$ & $73.92$ & \bm{$85.73$} & $91.07$ \\
\makecell{\textbf{RRFNet-56}\\(20\% mask)} & ResNet50 & Casia-WebFace & $99.55$ & $97.73$ & \bm{$91.13$} & \bm{$96.03$} & \bm{$94.47$} & $74.35$ & $84.93$ & $91.17$ \\
\makecell{\textbf{RRFNet-56}\\(w/o mask)} & ResNet50 & Casia-WebFace & \bm{$99.60$} & $97.86$ & $90.95$ & $95.65$ & $94.20$ & \bm{$75.42$} & $84.95$ & \bm{$91.23$} \\
\hline
AdaFace \cite{kim2022adaface} & ResNet100 & WebFace4M & $99.80$ & $99.26$ & $94.63$ & $97.90$ & $96.05$ & $84.50$ & $94.82$ & $95.28$ \\
ArcFace \cite{deng2019arcface} & ResNet100 & WebFace4M & $99.82$ & $99.27$ & $94.50$ & $98.02$ & $95.98$ & $84.35$ & $93.70$ & $95.09$ \\
KP-RPE \cite{vitkprpe} & ViT & WebFace4M & \bm{$99.83$} & $99.16$ & \bm{$95.40$} & $97.67$ & $96.00$ & $82.82$ & $90.67$ & $94.51$ \\
BagNet-17 \cite{brendel2018approximating} & ResNet100 & WebFace4M & $99.63$ & $97.89$ & $91.47$ & $95.50$ & $94.98$ & $75.33$ & $82.63$ & $91.06$ \\
BagNet-33 \cite{brendel2018approximating} & ResNet100 & WebFace4M & $99.67$ & $98.81$ & $93.28$ & $96.78$ & $95.62$ & $78.92$ & $86.15$ & $92.75$ \\

\makecell{\textbf{RRFNet-28}\\(40\% mask)} & ResNet100 & WebFace4M & $99.78$ & $99.23$ & $94.67$ & $98.03$ & $96.05$ & $82.85$ & $94.30$ & $94.99$ \\
\makecell{\textbf{RRFNet-28}\\(20\% mask)} & ResNet100 & WebFace4M & $99.73$ & $99.19$ & $94.75$ & $97.97$ & $95.88$ & $83.70$ & $95.17$ & $95.20$ \\
\makecell{\textbf{RRFNet-28}\\(w/o mask)} & ResNet100 & WebFace4M & $99.77$ & $99.30$ & $94.57$ & $98.02$ & $96.02$ & $83.37$ & $94.47$ & $95.07$ \\
\makecell{\textbf{RRFNet-56}\\(40\% mask)} & ResNet100 & WebFace4M & $99.83$ & \bm{$99.39$} & $95.25$ & \bm{$98.18$} & \bm{$96.08$} & $84.55$ & \bm{$96.52$} & \bm{$95.69$} \\
\makecell{\textbf{RRFNet-56}\\(20\% mask)} & ResNet100 & WebFace4M & $99.82$ & $99.36$ & $95.10$ & $98.18$ & $95.97$ & \bm{$84.58$} & $96.13$ & $95.59$ \\
\makecell{\textbf{RRFNet-56}\\(w/o mask)} & ResNet100 & WebFace4M & $99.75$ & $99.37$ & $94.98$ & $98.03$ & $95.97$ & $84.28$ & $95.80$ & $95.46$ \\
\hline
\end{tabular}
\label{tab:results_new}
\end{table*}

%% file: tables/cross_pose.tex
\begin{table}[bt]
\centering
\caption{Percentage of patches in genuine pairs whose most similar match occurs at the same spatial location.}
\label{tab:cross_pose_top_match}
\setlength{\tabcolsep}{2.5pt}
\resizebox{\columnwidth}{!}{%
\begin{tabular}{|l|c|c|c|c|c|c|c|}
\hline
Dataset & LFW & CFPFP & CPLFW & AGEDB & CALFW & ECL & HAD \\
\hline
Same region (\%) & $46.2$ & $20.3$ & $23.6$ & $34.6$ & $39.9$ & $48.9$ & $49.6$ \\
\hline
\end{tabular}%
}
\vspace{-10pt}
\end{table}

%% file: tables/patch56selection.tex
\begin{table}[b!]
\centering
\caption{Four patch configurations for $56 \times 56$ receptive fields. The configuration in the last row is chosen for the experiments as it achieves the highest accuracy.}
\begin{tabular}{|c|l|}
\hline
$\#$ Patches & Patch Positions (x, y) \\
\hline
4 & (0, 0), (0, 56), (56, 0), (56, 56) \\
4 & (0, 28), (28, 0), (56, 28), (28, 56) \\
5 & (0, 0), (0, 56), (56, 0), (56, 56), (28, 28) \\
\textbf{5} & \textbf{(0, 28), (28, 0), (56, 28), (28, 56), (28, 28)} \\

\hline
\end{tabular}
\label{tab:patchconf}
\end{table}

%% file: tables/inference_time.tex
\setlength{\tabcolsep}{4pt} 
\renewcommand{\arraystretch}{1.2}
\begin{table*}[hbt]
\centering
\caption{Comparison with explainable face verification approaches.}
\begin{tabular}{|c|c|c|c|c|c|c|c|c|c|}
\hline
\multirow{2}{*}{Method} & \multirow{2}{*}{\#\ Param.} & 
\multirow{2}{*}{\begin{tabular}[c]{@{}c@{}}Training\\Time (h)\end{tabular}} &
\multirow{2}{*}{\begin{tabular}[c]{@{}c@{}}Inference\\Time (s)\end{tabular}} &
\multicolumn{2}{c|}{LFW} & \multicolumn{2}{c|}{CFPFP} & \multicolumn{2}{c|}{CPLFW} \\
\cline{5-10}
& & & & Ins. $(\uparrow)$ & Del. $(\downarrow)$ & Ins. $(\uparrow)$ & Del. $(\downarrow)$ & Ins. $(\uparrow)$ & Del. $(\downarrow)$ \\
\hline
\hline
\begin{tabular}[c]{@{}c@{}}ArcFace \cite{deng2019arcface}\\(w/ random saliency)\end{tabular} & $65M$ & $17$ & $0.11$ &
\begin{tabular}[c]{@{}c@{}}$0.658$\\$\pm0.062$\end{tabular} & \begin{tabular}[c]{@{}c@{}}$0.658$\\$\pm0.060$\end{tabular} &
\begin{tabular}[c]{@{}c@{}}$0.601$\\$\pm0.084$\end{tabular} & \begin{tabular}[c]{@{}c@{}}$0.600$\\$\pm0.084$\end{tabular} &
\begin{tabular}[c]{@{}c@{}}$0.616$\\$\pm0.109$\end{tabular} & \begin{tabular}[c]{@{}c@{}}$0.617$\\$\pm0.109$\end{tabular} \\
\hline
\begin{tabular}[c]{@{}c@{}}ArcFace \cite{deng2019arcface}\\(w/ xSSAB \cite{huber2024efficient})\end{tabular} & $65M$ & $17$ & $0.52$ &
\begin{tabular}[c]{@{}c@{}}$0.660$\\$\pm0.071$\end{tabular} & \begin{tabular}[c]{@{}c@{}}$0.481$\\$\pm0.093$\end{tabular} &
\begin{tabular}[c]{@{}c@{}}$0.588$\\$\pm0.112$\end{tabular} & \begin{tabular}[c]{@{}c@{}}$0.476$\\$\pm0.121$\end{tabular} &
\begin{tabular}[c]{@{}c@{}}$0.635$\\$\pm0.129$\end{tabular} & \begin{tabular}[c]{@{}c@{}}$0.466$\\$\pm0.130$\end{tabular} \\
\hline
\begin{tabular}[c]{@{}c@{}}ArcFace \cite{deng2019arcface}\\(w/ xFace \cite{knoche2023explainable})\end{tabular} & $65M$ & $17$ & $61$ &
\begin{tabular}[c]{@{}c@{}}$0.787$\\$\pm0.042$\end{tabular} & \begin{tabular}[c]{@{}c@{}}$0.282$\\$\pm0.058$\end{tabular} &
\begin{tabular}[c]{@{}c@{}}$0.805$\\$\pm0.055$\end{tabular} & \begin{tabular}[c]{@{}c@{}}$0.262$\\$\pm0.092$\end{tabular} &
\begin{tabular}[c]{@{}c@{}}$0.761$\\$\pm0.101$\end{tabular} & \begin{tabular}[c]{@{}c@{}}$0.263$\\$\pm0.089$\end{tabular} \\
\hline
\begin{tabular}[c]{@{}c@{}}xCos \cite{xcos}\\(Self-interpretable)\end{tabular} & $103M$ & 12 & $0.09$ &
\begin{tabular}[c]{@{}c@{}}$0.647$\\$\pm0.071$\end{tabular} & \begin{tabular}[c]{@{}c@{}}$0.250$\\$\pm0.060$\end{tabular} &
\begin{tabular}[c]{@{}c@{}}$0.532$\\$\pm0.118$\end{tabular} & \begin{tabular}[c]{@{}c@{}}$0.354$\\$\pm0.136$\end{tabular} &
\begin{tabular}[c]{@{}c@{}}$0.544$\\$\pm0.122$\end{tabular} & \begin{tabular}[c]{@{}c@{}}$0.348$\\$\pm0.137$\end{tabular} \\
\hline
\begin{tabular}[c]{@{}c@{}}\textbf{RRFNet-28}\\(Self-interpretable)\end{tabular} & $56M$ & $81$ & $0.29$ &
\begin{tabular}[c]{@{}c@{}}\bm{$0.822$}\\$\pm0.028$\end{tabular} & \begin{tabular}[c]{@{}c@{}}\bm{$0.199$}\\$\pm0.032$\end{tabular} &
\begin{tabular}[c]{@{}c@{}}\bm{$0.852$}\\$\pm0.031$\end{tabular} & \begin{tabular}[c]{@{}c@{}}\bm{$0.166$}\\$\pm0.043$\end{tabular} &
\begin{tabular}[c]{@{}c@{}}\bm{$0.829$}\\$\pm0.069$\end{tabular} & \begin{tabular}[c]{@{}c@{}}\bm{$0.186$}\\$\pm0.072$\end{tabular} \\
\hline
\begin{tabular}[c]{@{}c@{}}\textbf{RRFNet-56}\\(Self-interpretable)\end{tabular} & $65M$ & $42$ & $0.21$ &
\begin{tabular}[c]{@{}c@{}}$0.819$\\$\pm0.031$\end{tabular} & \begin{tabular}[c]{@{}c@{}}$0.223$\\$\pm0.035$\end{tabular} &
\begin{tabular}[c]{@{}c@{}}$0.829$\\$\pm0.036$\end{tabular} & \begin{tabular}[c]{@{}c@{}}$0.208$\\$\pm0.044$\end{tabular} &
\begin{tabular}[c]{@{}c@{}}$0.797$\\$\pm0.095$\end{tabular} & \begin{tabular}[c]{@{}c@{}}$0.234$\\$\pm0.087$\end{tabular} \\
\hline
\end{tabular}
\label{tab:comparison_previous_work}
\end{table*}

%% file: figures/del_imp_gen_fig.tex
\begin{figure}[htb]
    \centering
    \setlength{\tabcolsep}{1pt}
    
    \hspace{-4pt}\makebox[0.185\linewidth][c]{\footnotesize Reference}\hspace{8pt}%
    \makebox[0.74\linewidth][c]{\footnotesize Probe}\\[2pt]
    
    \begin{tabular}{ccccc}
        & \hspace{2pt}
        \makebox[0.185\linewidth][c]{\footnotesize 0\% mask} &
        \makebox[0.185\linewidth][c]{\footnotesize 5\% mask} &
        \makebox[0.185\linewidth][c]{\footnotesize 10\% mask} &
        \makebox[0.185\linewidth][c]{\footnotesize 15\% mask} \\
        \includegraphics[width=0.185\linewidth]{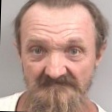} & \hspace{2pt}
        \includegraphics[width=0.185\linewidth]{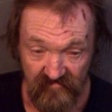} &
        \includegraphics[width=0.185\linewidth]{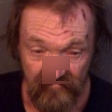} &
        \includegraphics[width=0.185\linewidth]{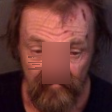} &
        \includegraphics[width=0.185\linewidth]{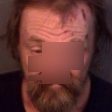} \\
        [-8pt]
         & \hspace{2pt}
        \makebox[0.185\linewidth][c]{\footnotesize score=0.62} &
        \makebox[0.185\linewidth][c]{\footnotesize score=0.49} &
        \makebox[0.185\linewidth][c]{\footnotesize score=0.33} &
        \makebox[0.185\linewidth][c]{\footnotesize score=0.21} \\
        \includegraphics[width=0.185\linewidth]{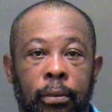} & \hspace{2pt}
        \includegraphics[width=0.185\linewidth]{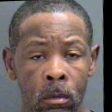} &
        \includegraphics[width=0.185\linewidth]{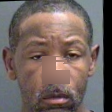} &
        \includegraphics[width=0.185\linewidth]{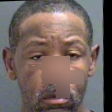} &
        \includegraphics[width=0.185\linewidth]{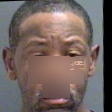} \\
        [-8pt]
         & \hspace{2pt}
        \makebox[0.185\linewidth][c]{\footnotesize score=0.45} &
        \makebox[0.185\linewidth][c]{\footnotesize score=0.24} &
        \makebox[0.185\linewidth][c]{\footnotesize score=0.21} &
        \makebox[0.185\linewidth][c]{\footnotesize score=0.18} \\
        \includegraphics[width=0.185\linewidth]{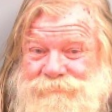} & \hspace{2pt}
        \includegraphics[width=0.185\linewidth]{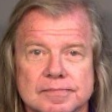} &
        \includegraphics[width=0.185\linewidth]{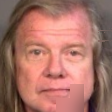} &
        \includegraphics[width=0.185\linewidth]{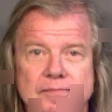} &
        \includegraphics[width=0.185\linewidth]{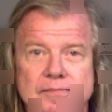} \\
        [-8pt]
         & \hspace{2pt}
        \makebox[0.185\linewidth][c]{\footnotesize score=0.44} &
        \makebox[0.185\linewidth][c]{\footnotesize score=0.48} &
        \makebox[0.185\linewidth][c]{\footnotesize score=0.49} &
        \makebox[0.185\linewidth][c]{\footnotesize score=0.48} \\
        \includegraphics[width=0.185\linewidth]{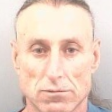} & \hspace{2pt}
        \includegraphics[width=0.185\linewidth]{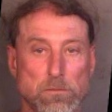} &
        \includegraphics[width=0.185\linewidth]{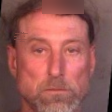} &
        \includegraphics[width=0.185\linewidth]{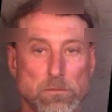} &
        \includegraphics[width=0.185\linewidth]{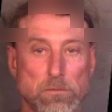} \\
        [-8pt]
         & \hspace{2pt}
        \makebox[0.185\linewidth][c]{\footnotesize score=0.43} &
        \makebox[0.185\linewidth][c]{\footnotesize score=0.45} &
        \makebox[0.185\linewidth][c]{\footnotesize score=0.50} &
        \makebox[0.185\linewidth][c]{\footnotesize score=0.50} \\
    \end{tabular}
    
    \caption{Examples of impostor (top two) and genuine (bottom two) pairs on Hadrian dataset. Similarity scores are given below each probe image. For impostor pairs, salient regions are masked, while for genuine pairs least salient regions are masked.}
    \label{fig:deletion_examples}
\end{figure}

%% file: figures/fpr_and_fnr.tex
\begin{figure}[htb]
    \centering
    \includegraphics[width=0.9\columnwidth]{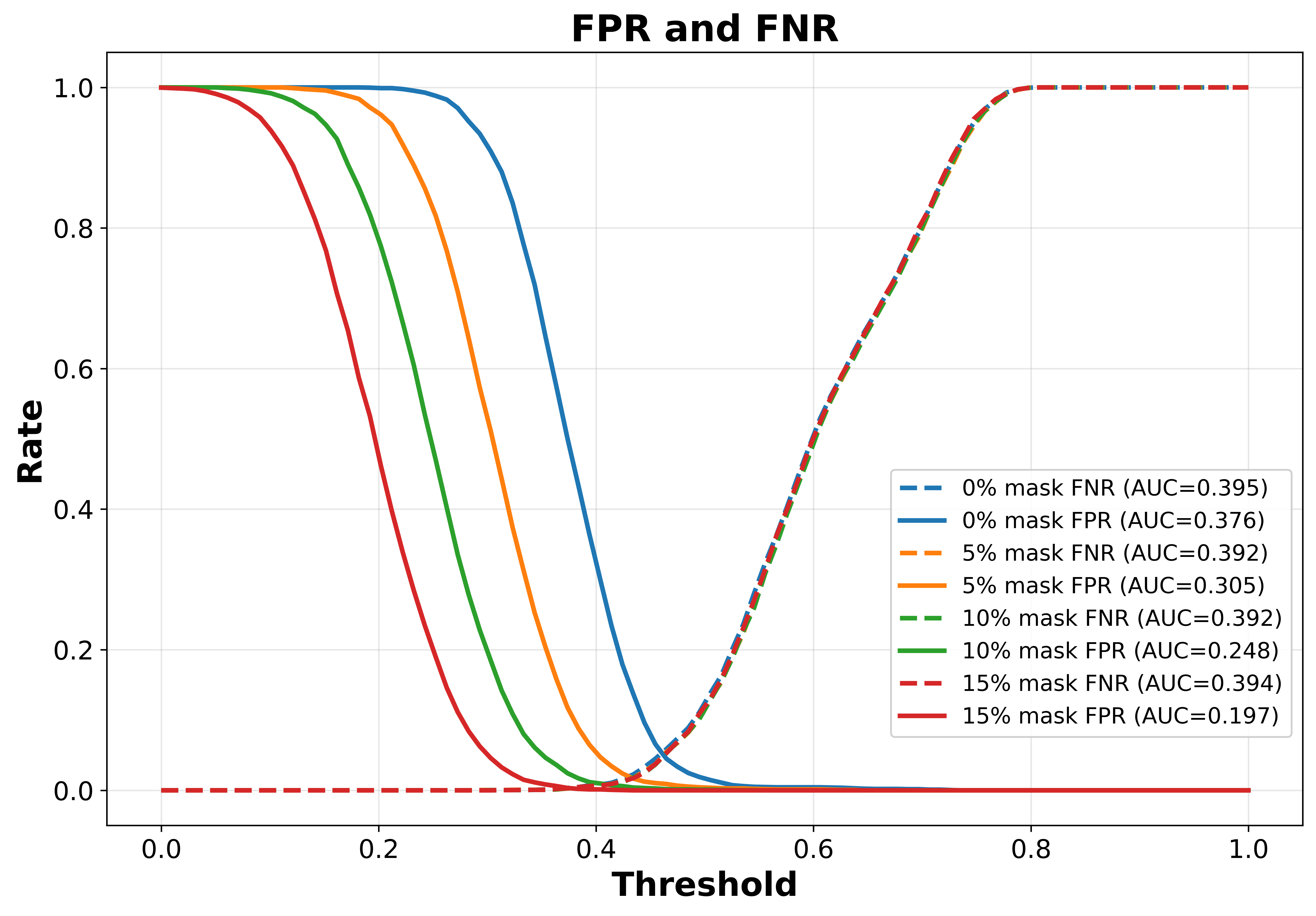}
    \caption{FPR and FNR analysis on Hadrian dataset.}
    \label{fig:fpr_and_fnr}
\end{figure}

%% file: sec/5_quantative_analysis.tex
\section{Analysis on Explainability}
\label{sec:quantative}
\input{figures/ins_del}

In this section, we compare our approach with prior works on explainability methods for face verification. Two post-hoc approaches, xSSAB \cite{huber2024efficient} and xFace \cite{knoche2023explainable}, and a self-interpretable method, xCos \cite{xcos}, are used for the comparison. For post-hoc methods, ArcFace \cite{deng2019arcface} is used as the backbone network.

\textbf{Evaluation Metrics.} Faithfulness of the generated saliency maps is measured using the Insertion and Deletion metrics \cite{rise}. The Insertion Area Under the Curve (AUC) score quantifies the increase in similarity scores as salient facial regions are progressively revealed, whereas the Deletion AUC score measures the decrease in similarity as salient regions are removed. In addition, the False Positive Rate (FPR) and False Negative Rate (FNR) are used to analyze the impact of masking the most salient region in impostor pairs and the least salient region in genuine pairs.

Comparison with previous approaches is presented in Table~\ref{tab:comparison_previous_work}. Genuine pairs from the LFW, CFPFP, and CPLFW datasets are used to report Insertion and Deletion scores for evaluating the quality of saliency maps. Training time is measured using four L40S GPUs, and inference time is reported for a pair of images on a CPU. We observe that, our approach outperforms previous works in both Insertion and Deletion metrics. Although xFace \cite{knoche2023explainable} achieves scores closest to ours, its inference time is substantially higher, as it requires 2,260 forward passes to generate a single saliency map. Examples for comparison between RRFNet-28 and xFace are shown in Fig. \ref{fig:rrfnet_vs_xface}. In some cases, the Insertion and Deletion scores of xFace fall to the level of random chance, highlighting the vulnerability of perturbation-based post-hoc methods.

Next, we analyze the effect of masking salient and least salient regions on impostor and genuine pairs, respectively. FPR and FNR curves for different masking rates are shown in Fig. \ref{fig:fpr_and_fnr}, and examples are presented in Fig. \ref{fig:deletion_examples}. Masking is performed by applying Gaussian blurring. We observe that the most salient regions typically correspond to the nose and eye areas, and masking these regions leads to a sharp decrease in similarity scores for impostors. Note, while a similar impact is also observed for genuine pairs, we investigate the impact of masking least salient regions to observe whether such masking can increase similarity scores. Our analysis shows that, while some pairs exhibit increased similarity after masking, typically on hair region, the overall FNR remains unchanged.

%% file: figures/ins_del.tex
\begin{figure*}[htbp]
    \centering
    
    \begin{minipage}{1\textwidth}
        \makebox[0.74\textwidth]{\textbf{RRFNet-28}}
        \makebox[0.11\textwidth]{\textbf{xFace}}
    \end{minipage}
    
    \vspace{0.4em}

    \begin{minipage}{1\textwidth}
        \centering
        \includegraphics[width=0.13\textwidth]{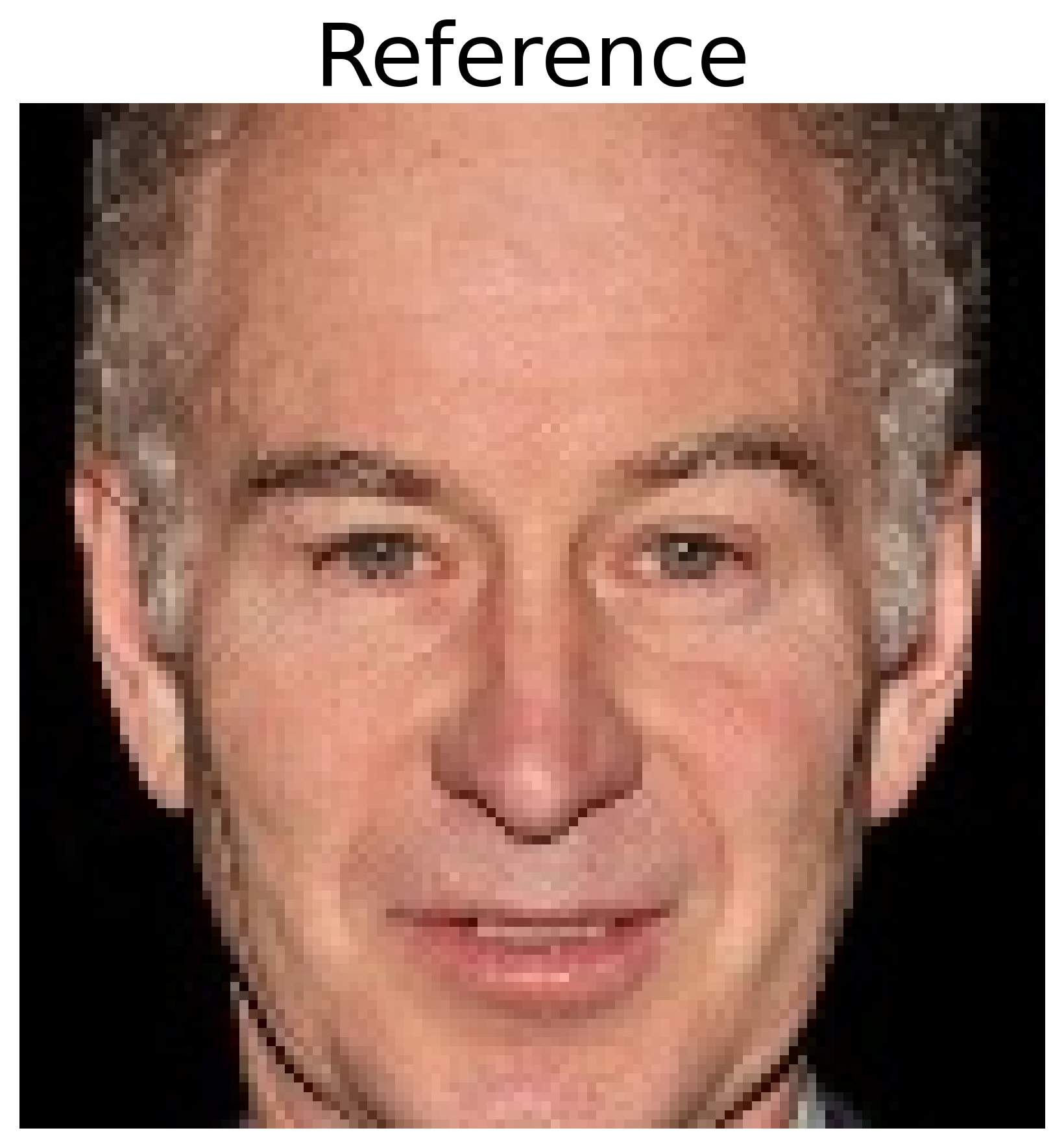}
        \raisebox{0\height}{\tikz\draw[dashed, line width=0.4pt] (0,0) -- (0,0.13\textwidth);}%
        \hspace{0.5em}
        \includegraphics[width=0.13\textwidth]{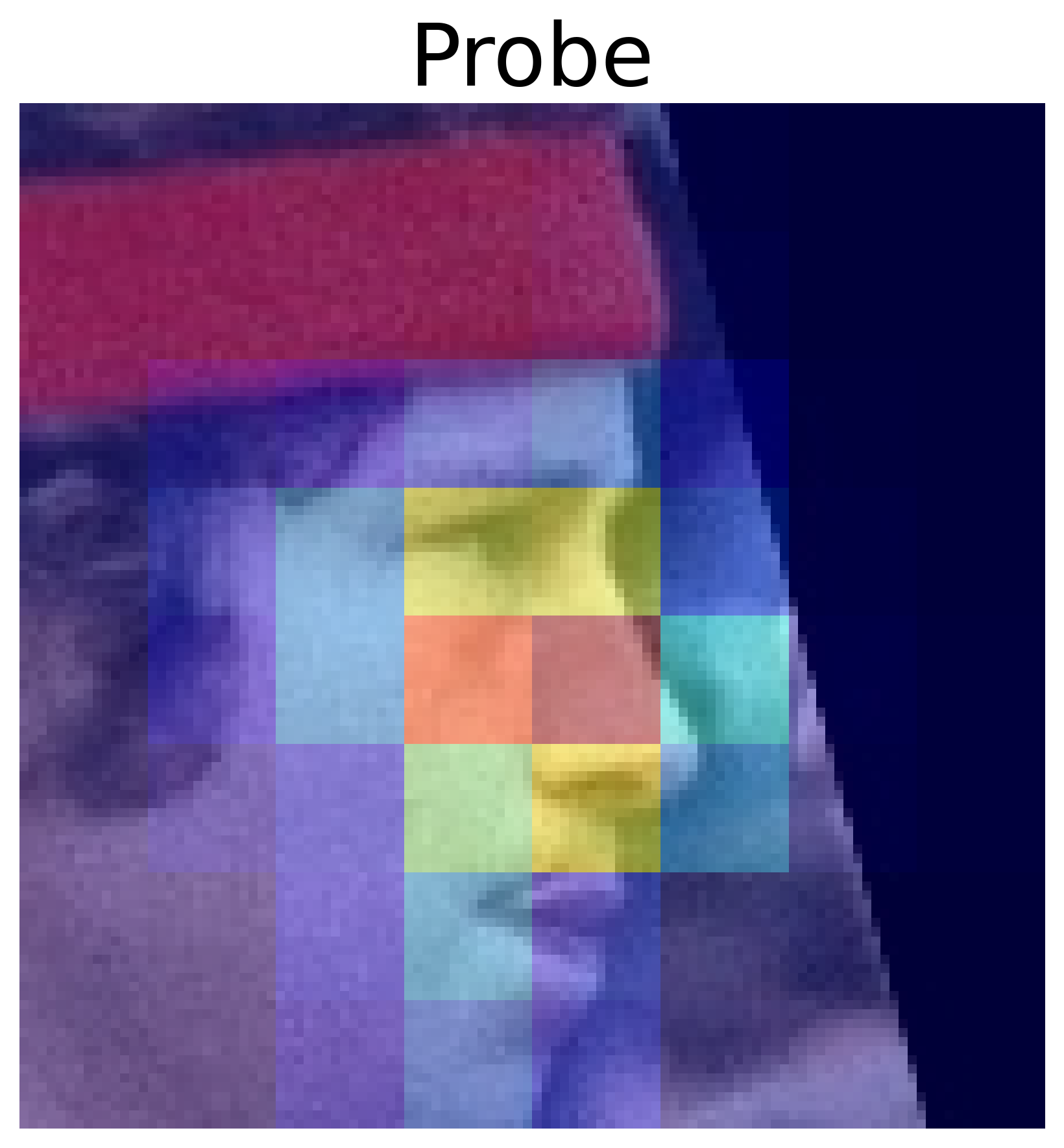}
        \includegraphics[width=0.13\textwidth]{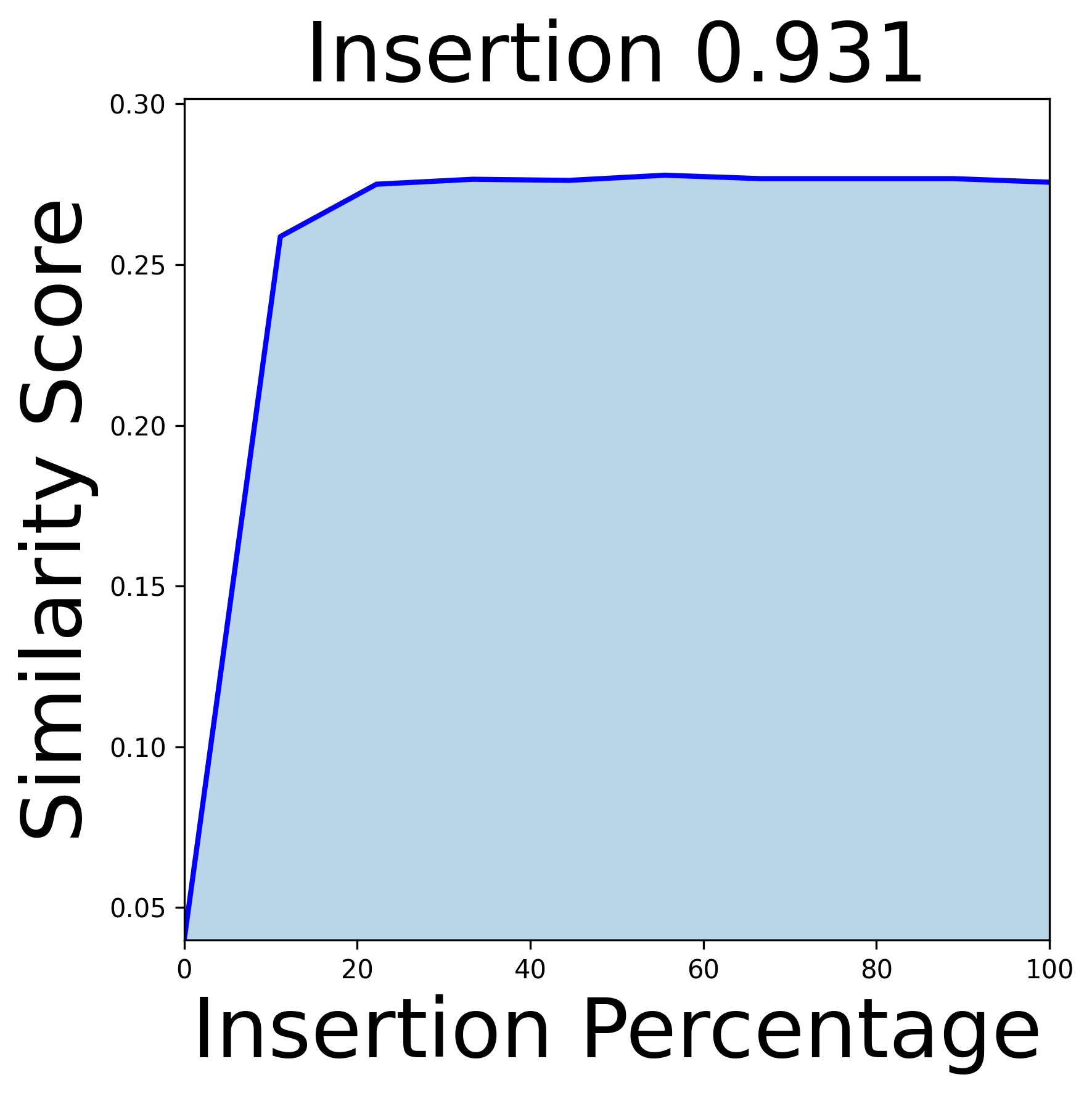}
        \includegraphics[width=0.13\textwidth]{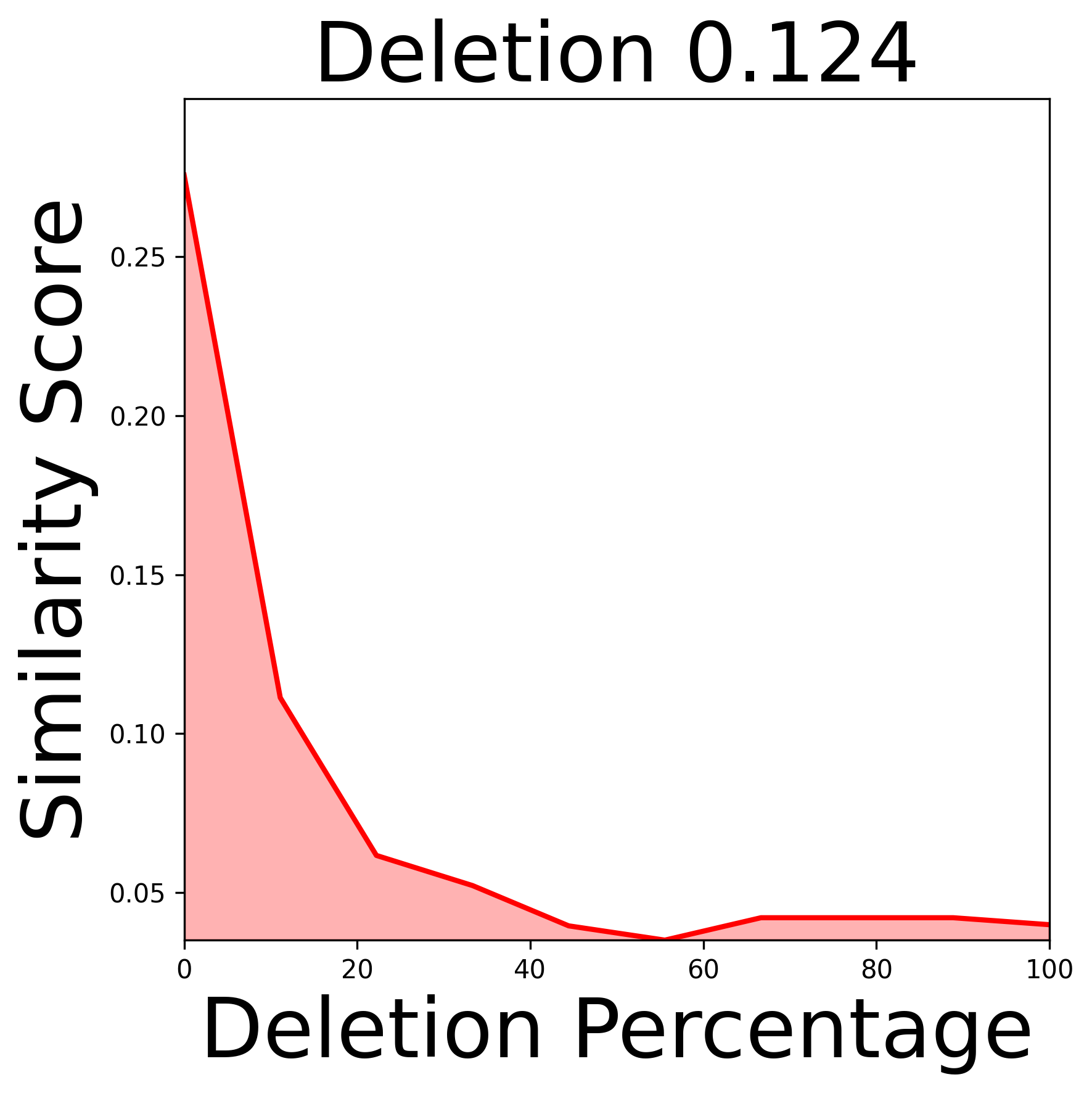}
        \raisebox{0\height}{\tikz\draw[dashed, line width=0.4pt] (0,0) -- (0,0.13\textwidth);}%
        \hspace{0.5em}
        \includegraphics[width=0.13\textwidth]{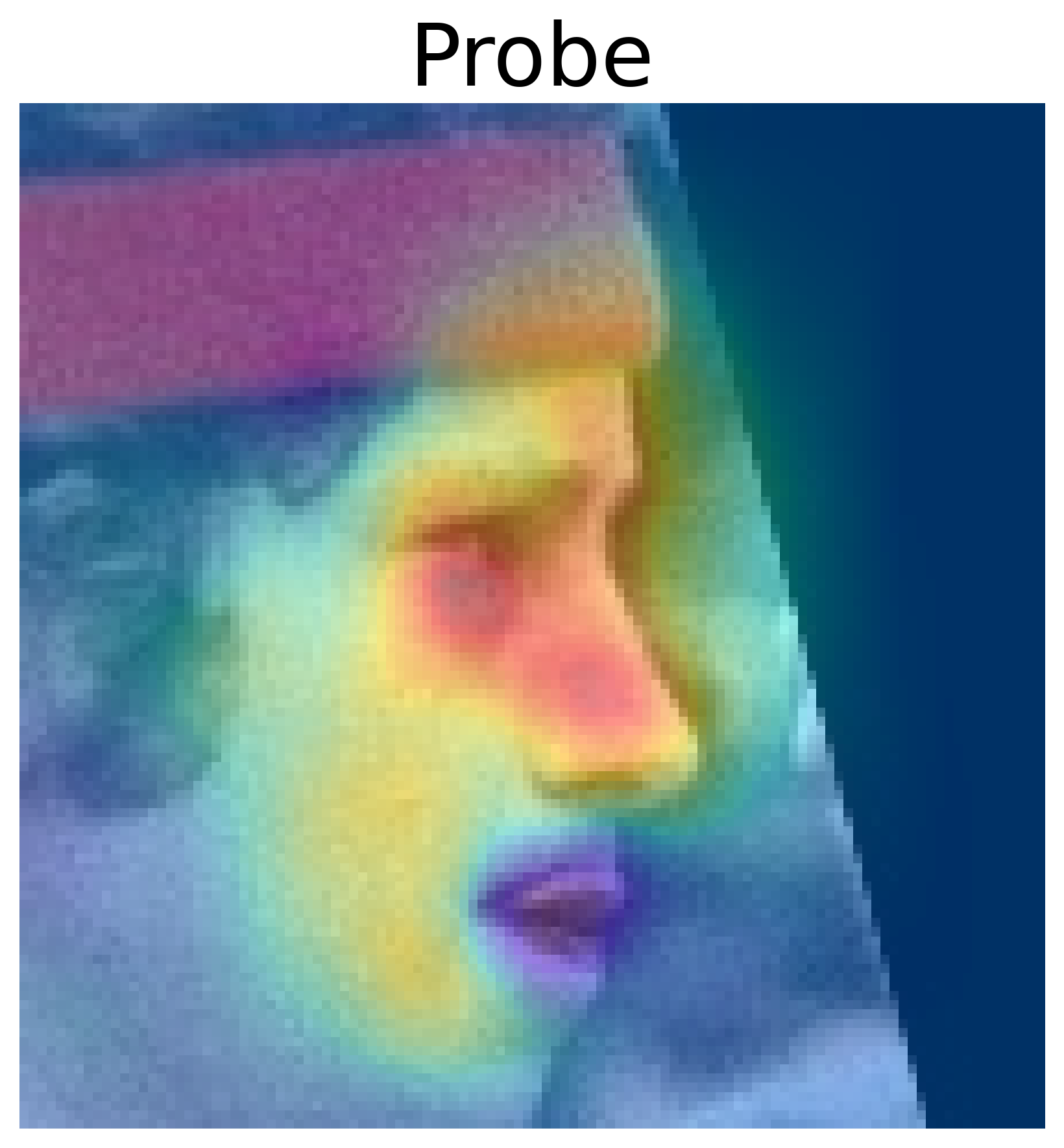}
        \includegraphics[width=0.13\textwidth]{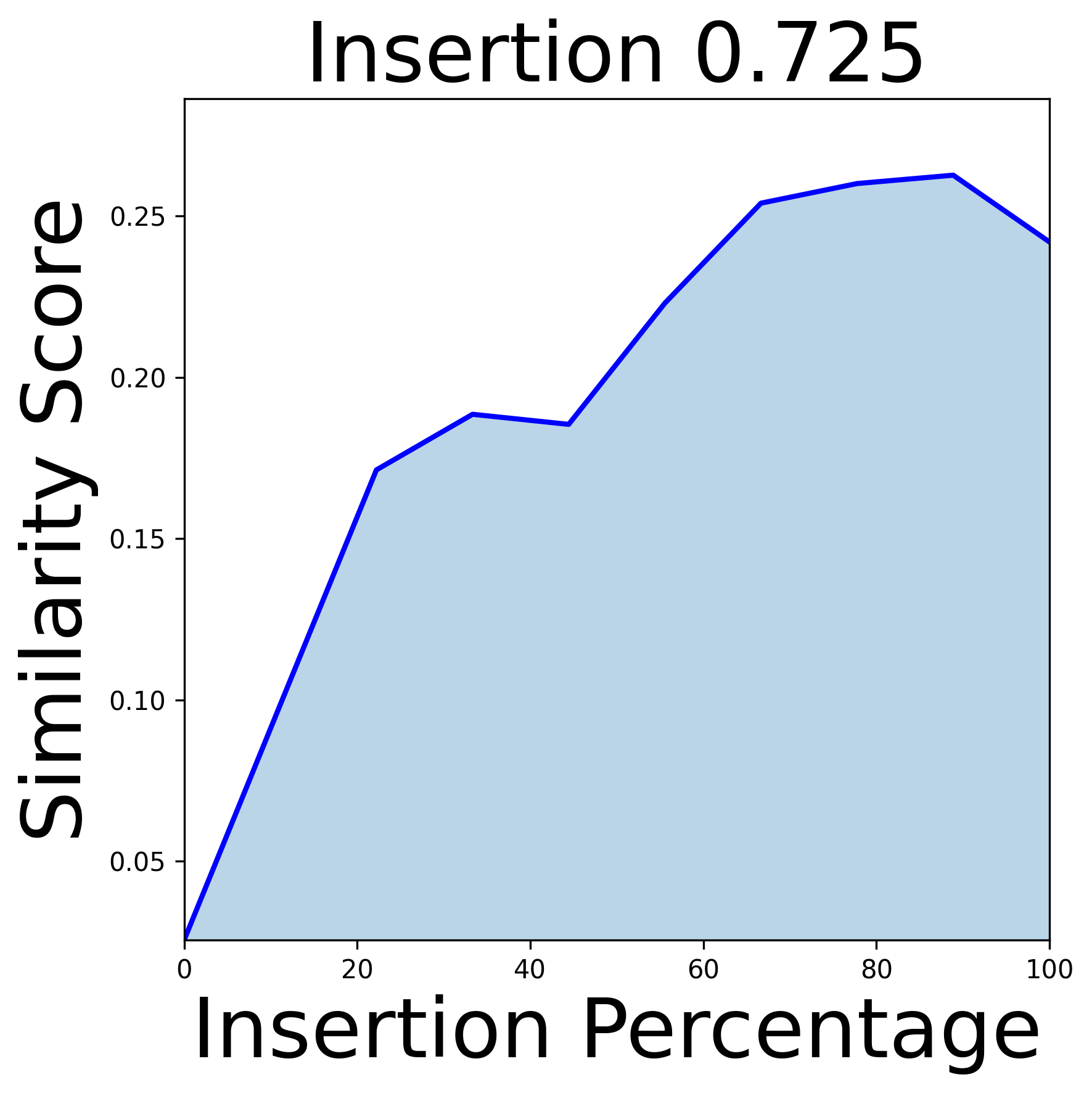}
        \includegraphics[width=0.13\textwidth]{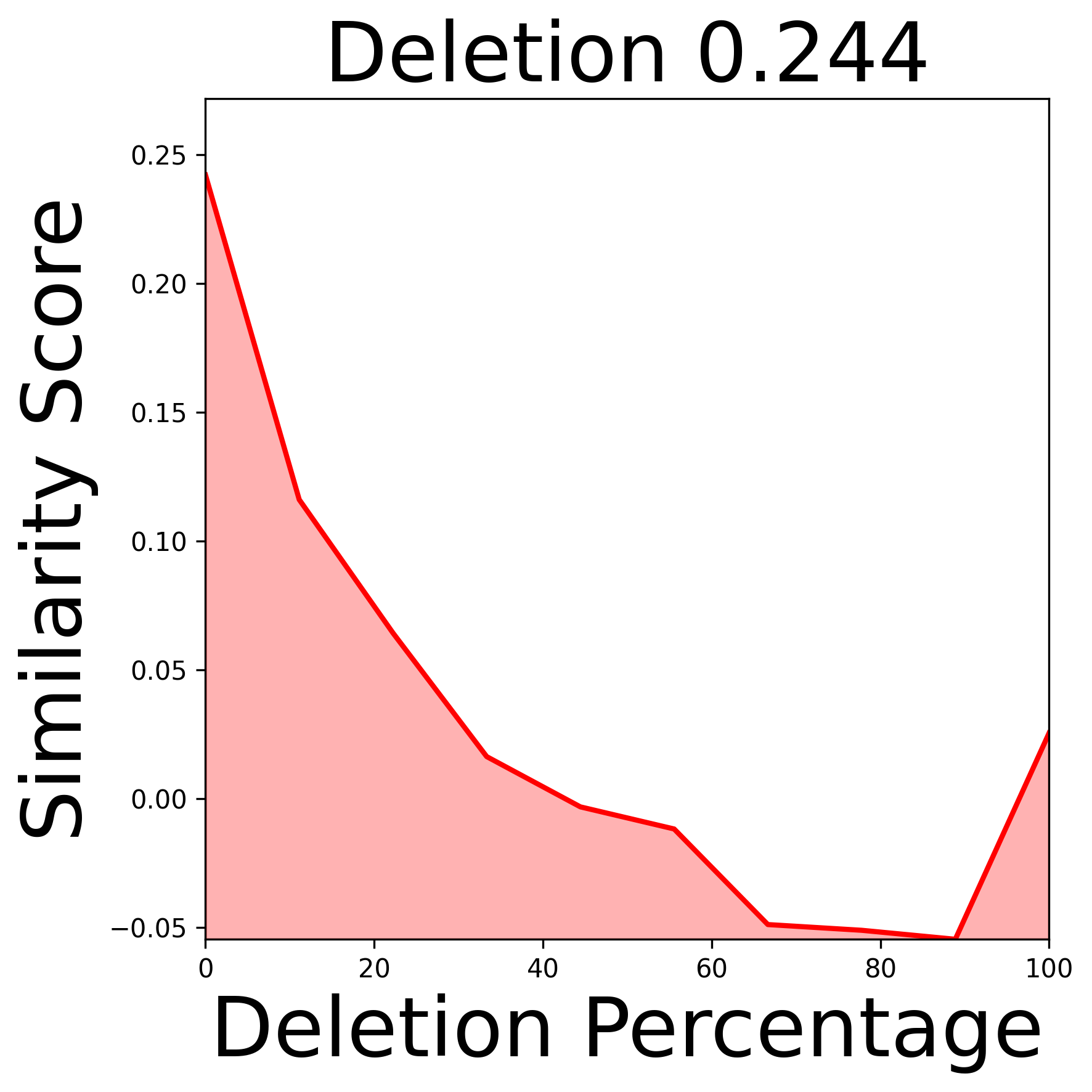}
    \end{minipage}
    
    \vspace{0.5em}

    \begin{minipage}{1\textwidth}
        \centering
        \includegraphics[width=0.13\textwidth]{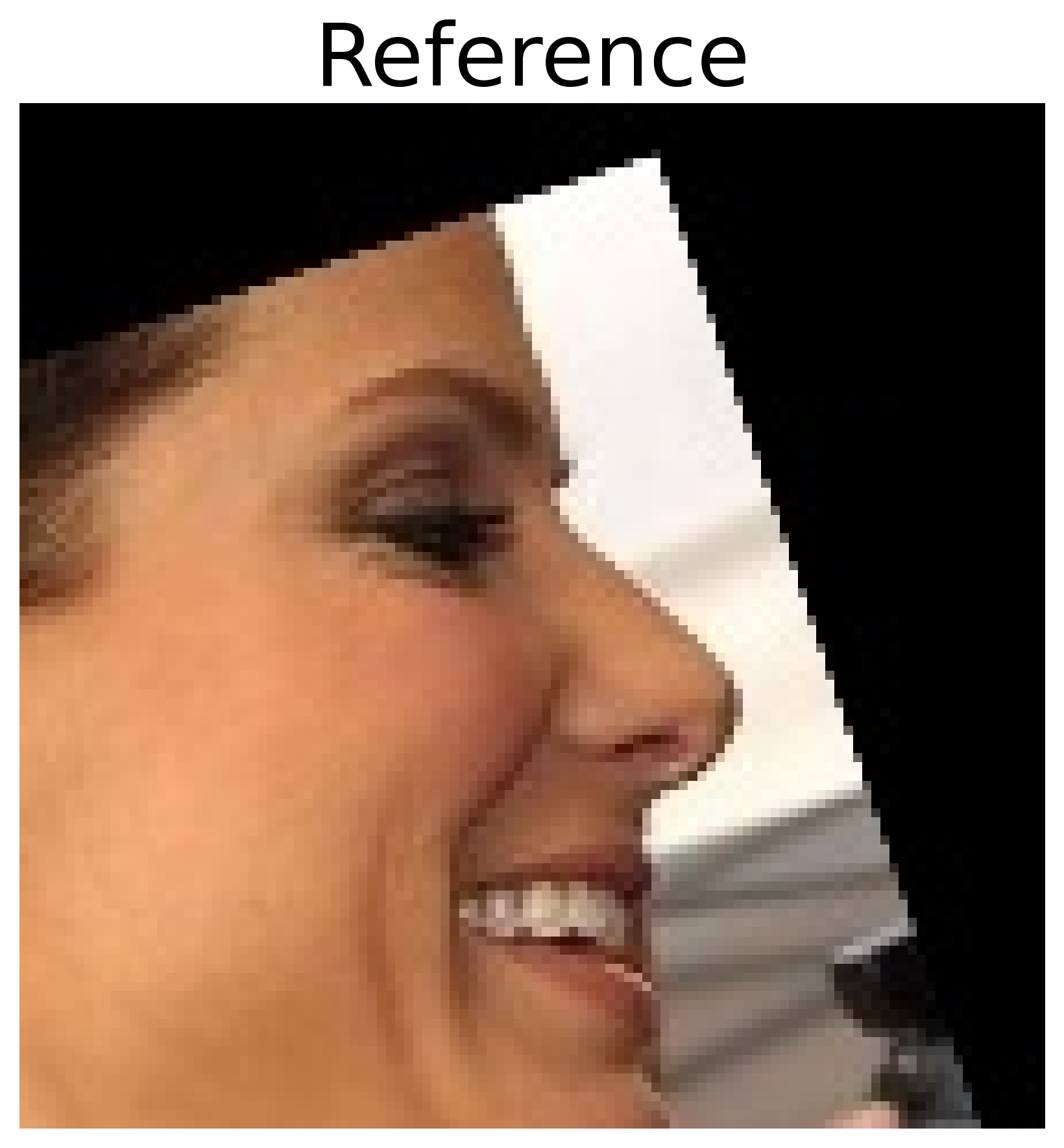}
        \raisebox{0\height}{\tikz\draw[dashed, line width=0.4pt] (0,0) -- (0,0.13\textwidth);}%
        \hspace{0.5em}
        \includegraphics[width=0.13\textwidth]{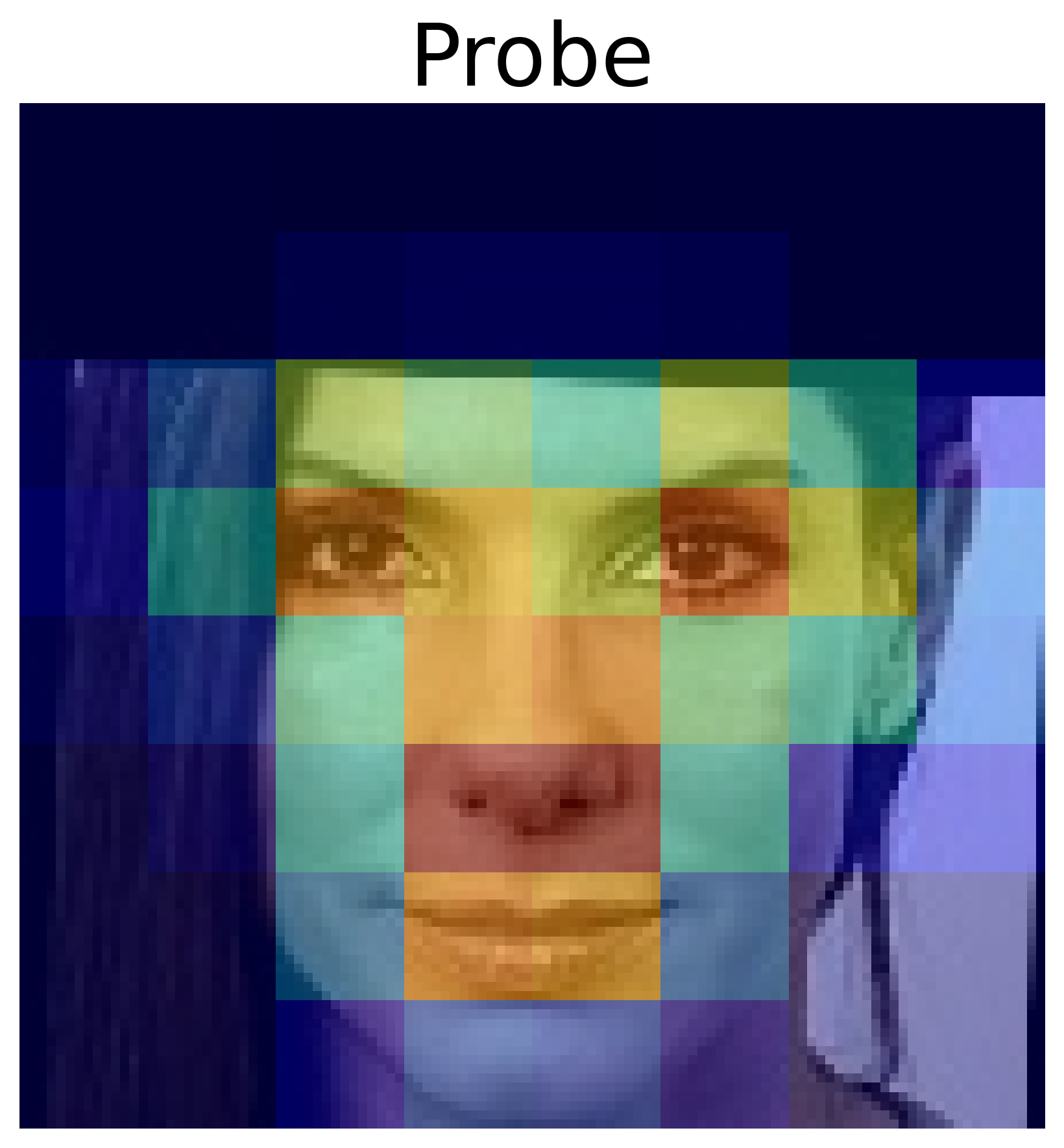}
        \includegraphics[width=0.13\textwidth]{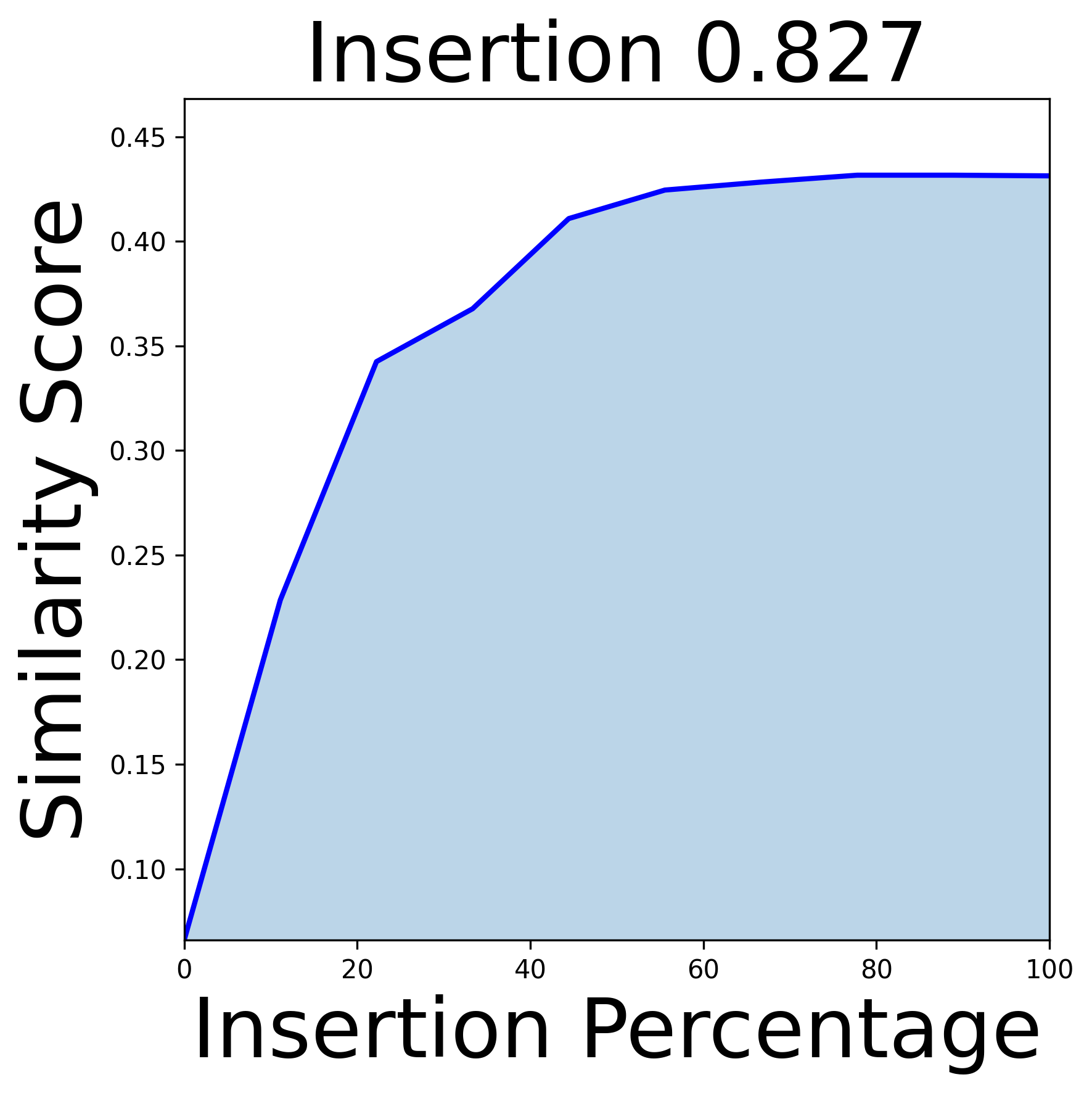}
        \includegraphics[width=0.13\textwidth]{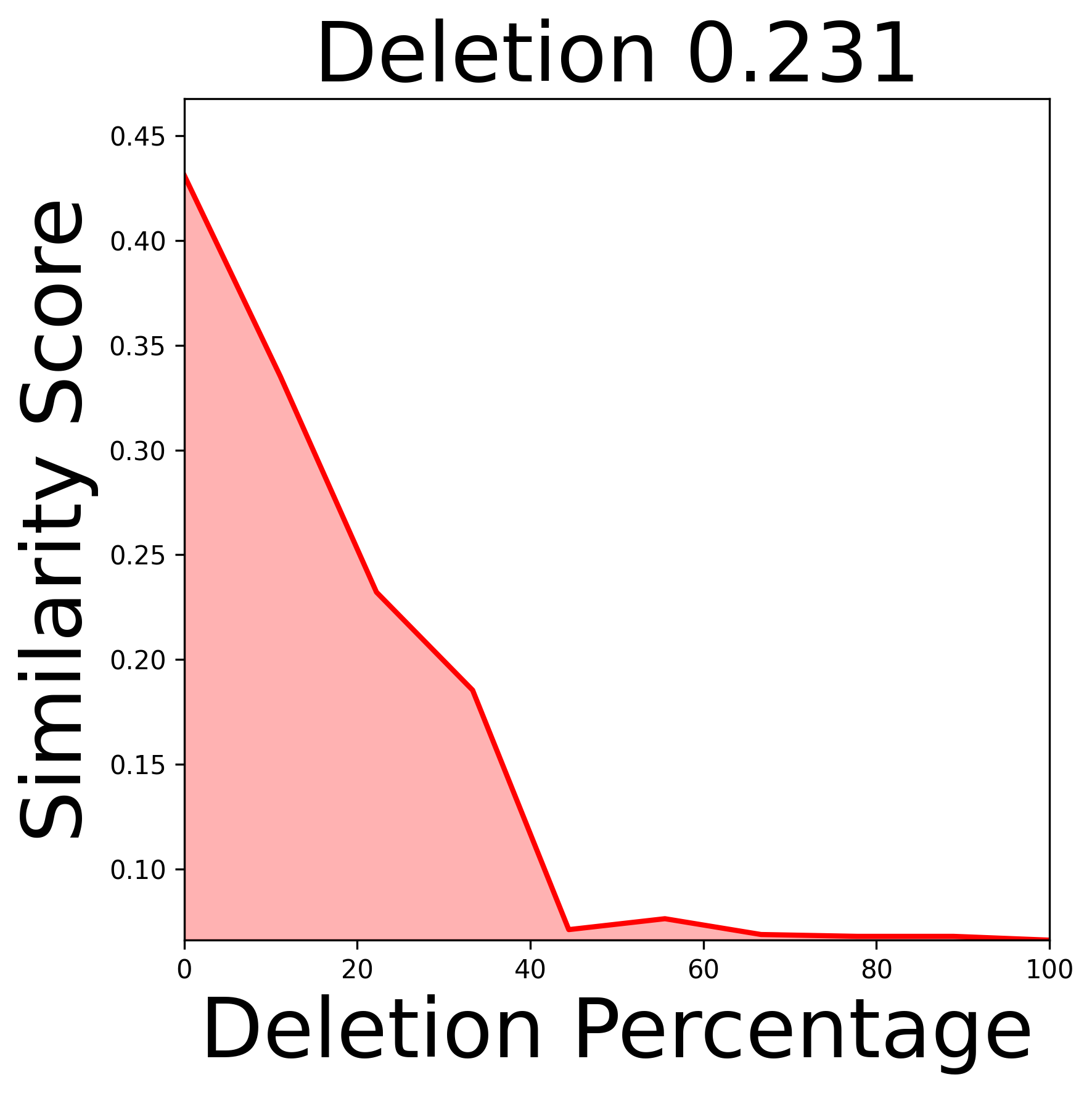}
        \raisebox{0\height}{\tikz\draw[dashed, line width=0.4pt] (0,0) -- (0,0.13\textwidth);}%
        \hspace{0.5em}
        \includegraphics[width=0.13\textwidth]{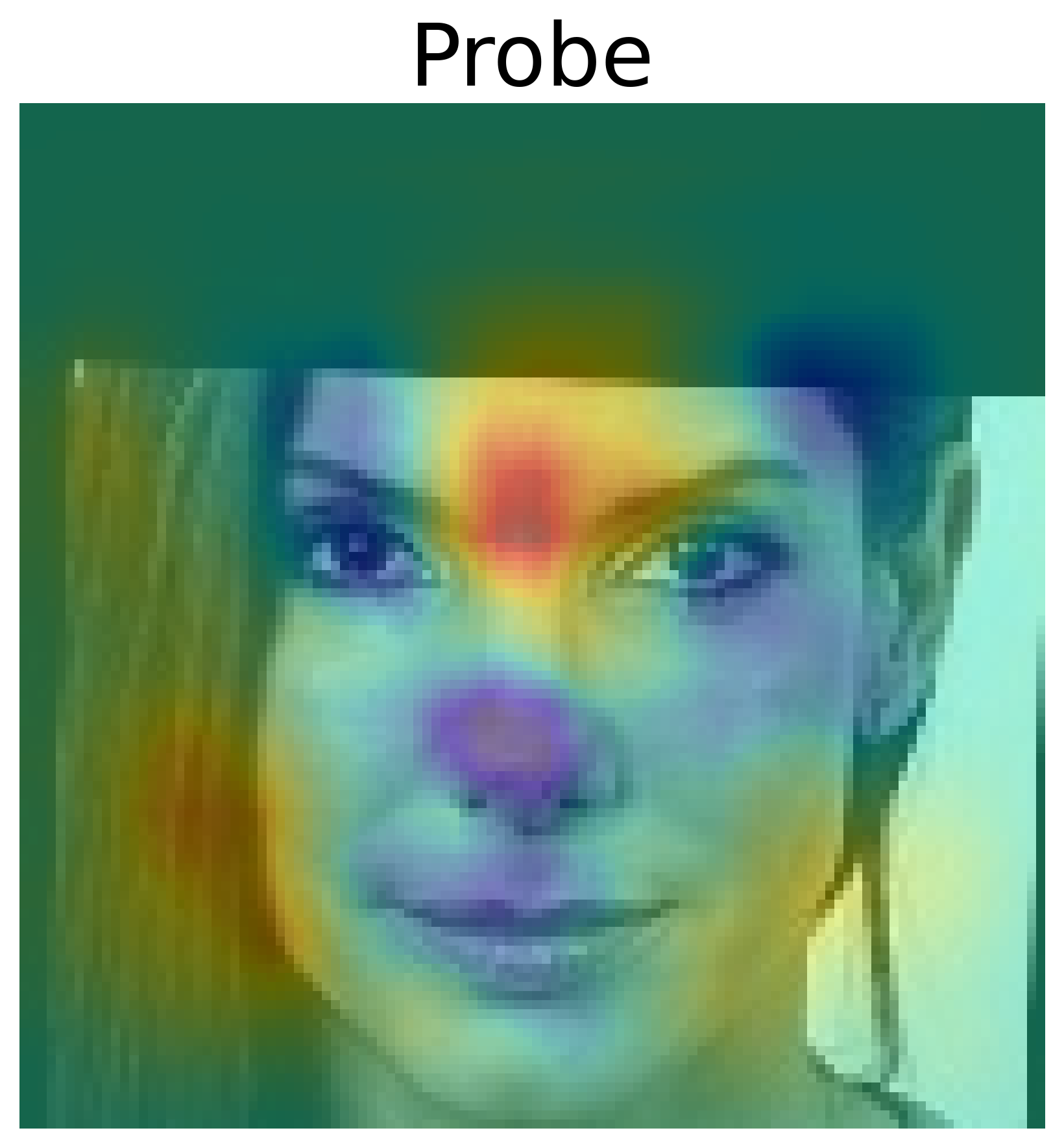}
        \includegraphics[width=0.13\textwidth]{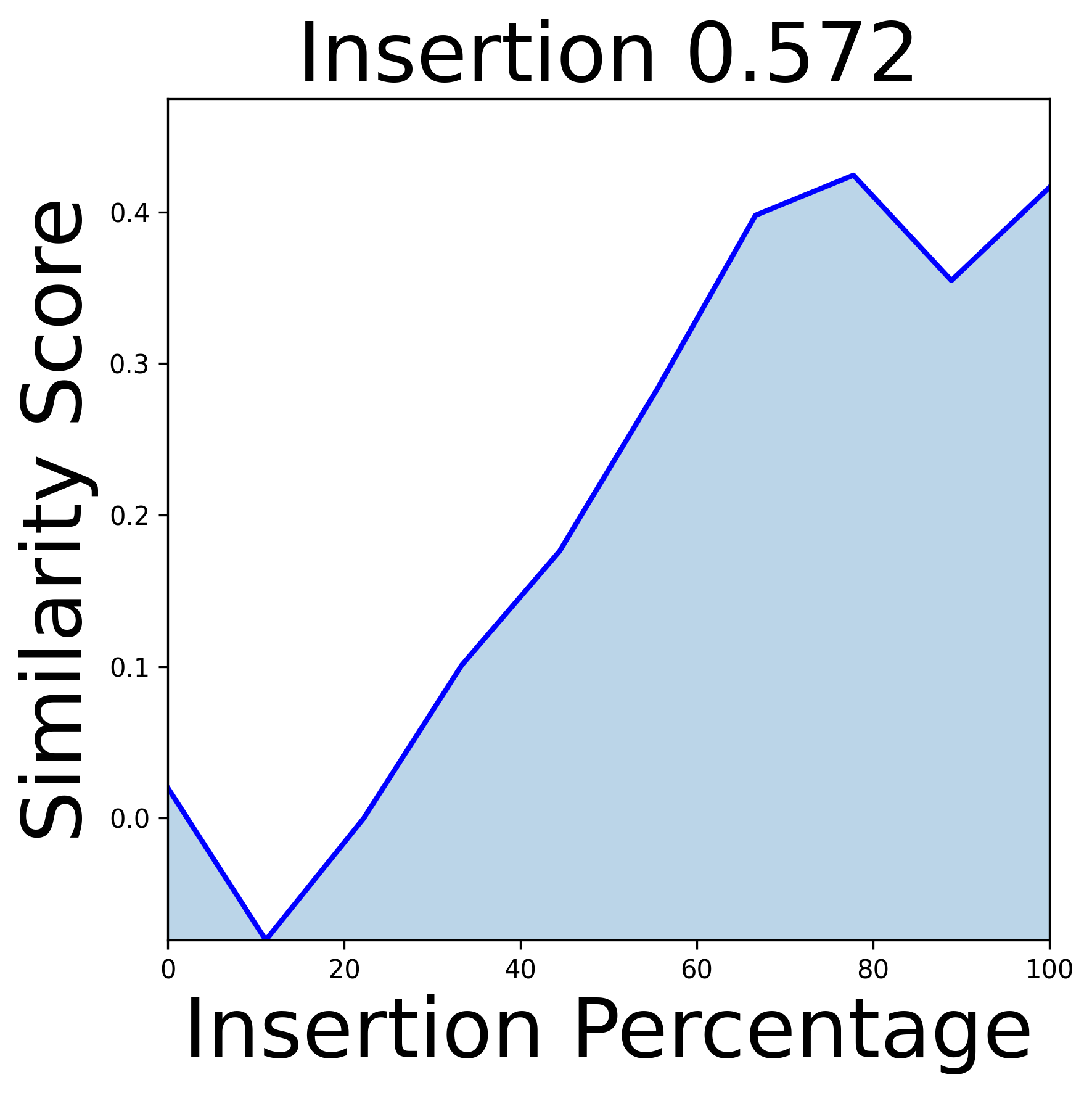}
        \includegraphics[width=0.13\textwidth]{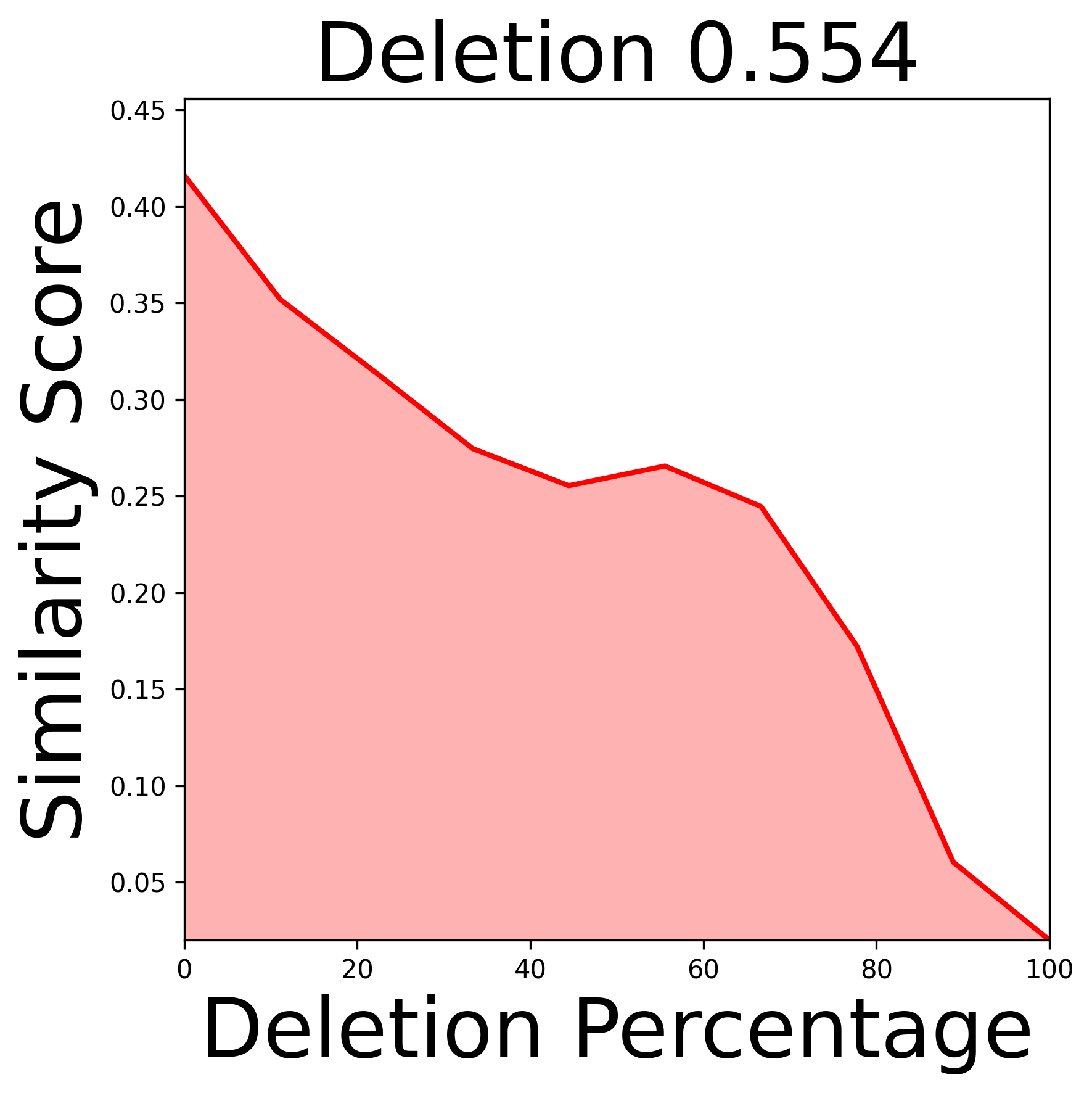}
    \end{minipage}
    
    \vspace{0.5em}

    \begin{minipage}{1\textwidth}
        \centering
        \includegraphics[width=0.13\textwidth]{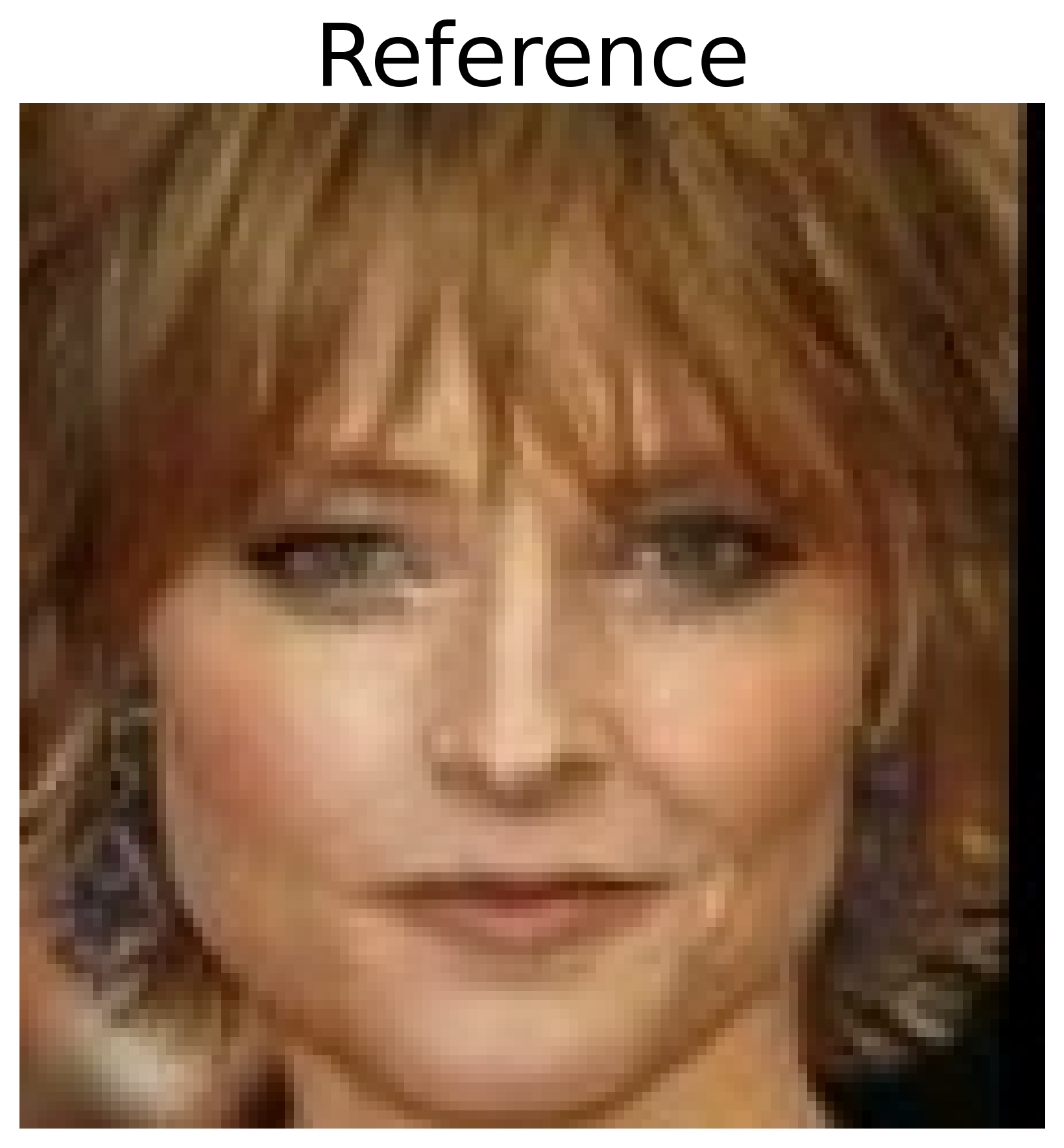}
        \raisebox{0\height}{\tikz\draw[dashed, line width=0.4pt] (0,0) -- (0,0.13\textwidth);}%
        \hspace{0.5em}
        \includegraphics[width=0.13\textwidth]{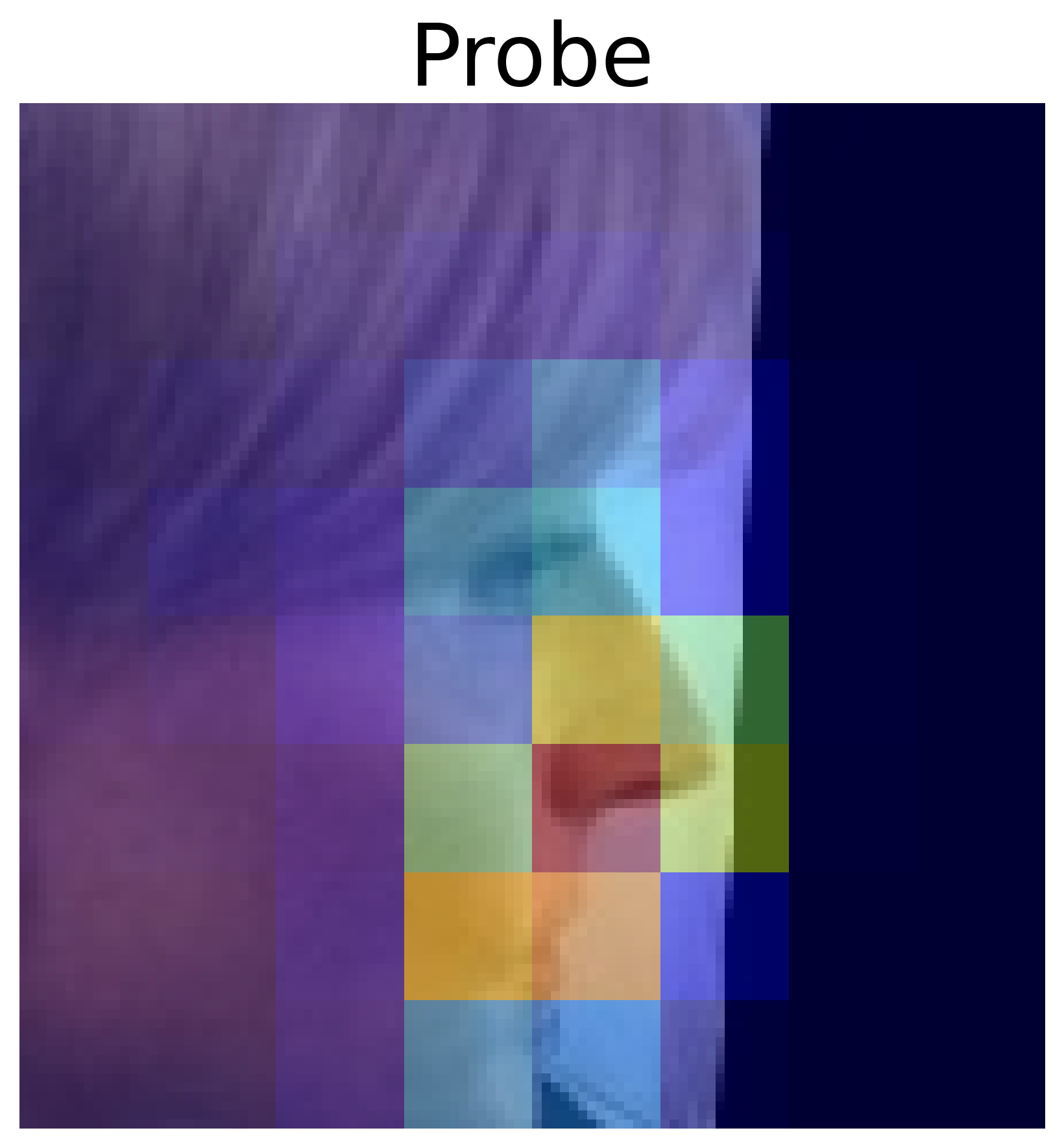}
        \includegraphics[width=0.13\textwidth]{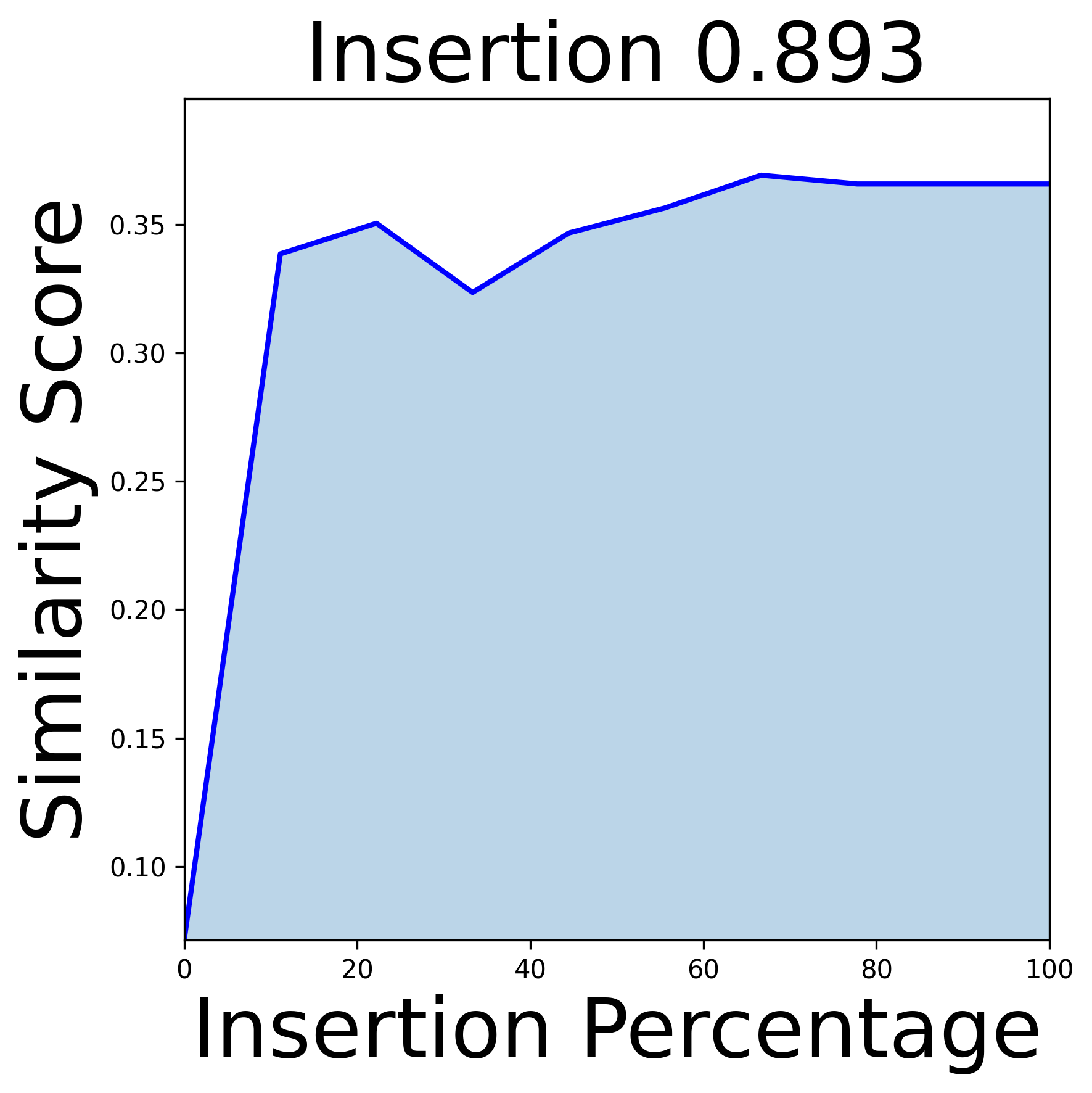}
        \includegraphics[width=0.13\textwidth]{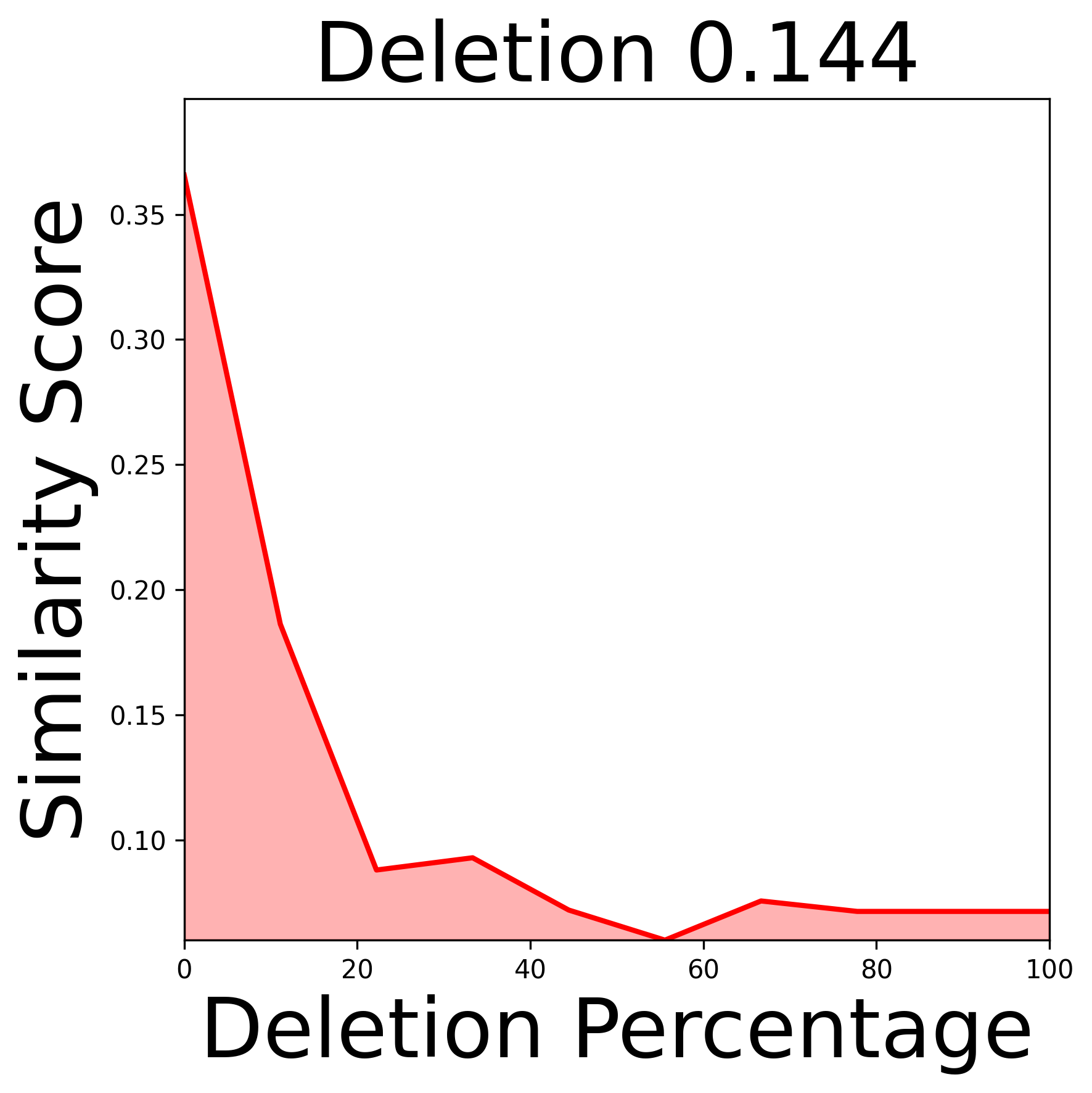}
        \raisebox{0\height}{\tikz\draw[dashed, line width=0.4pt] (0,0) -- (0,0.13\textwidth);}%
        \hspace{0.5em}
        \includegraphics[width=0.13\textwidth]{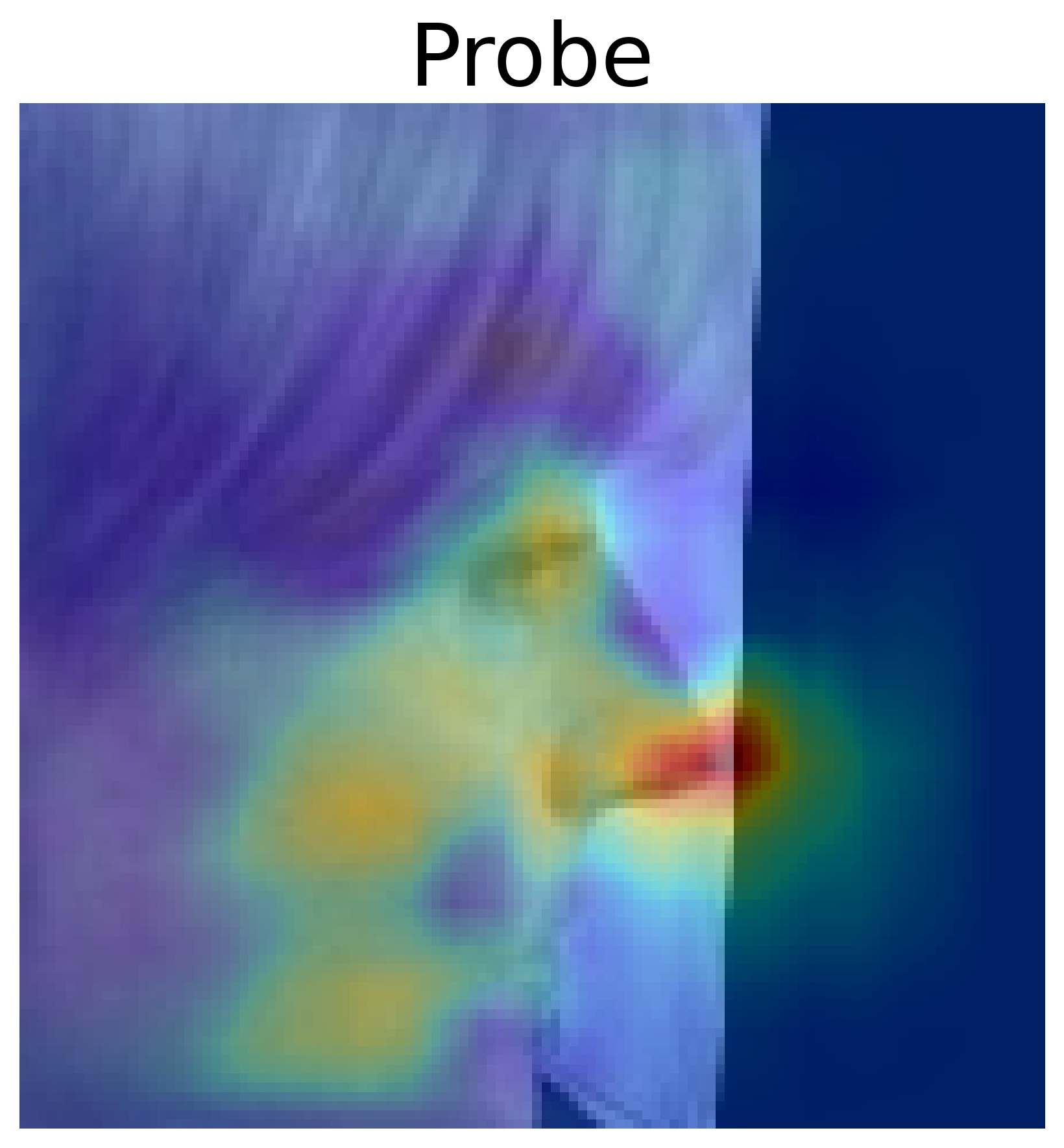}
        \includegraphics[width=0.13\textwidth]{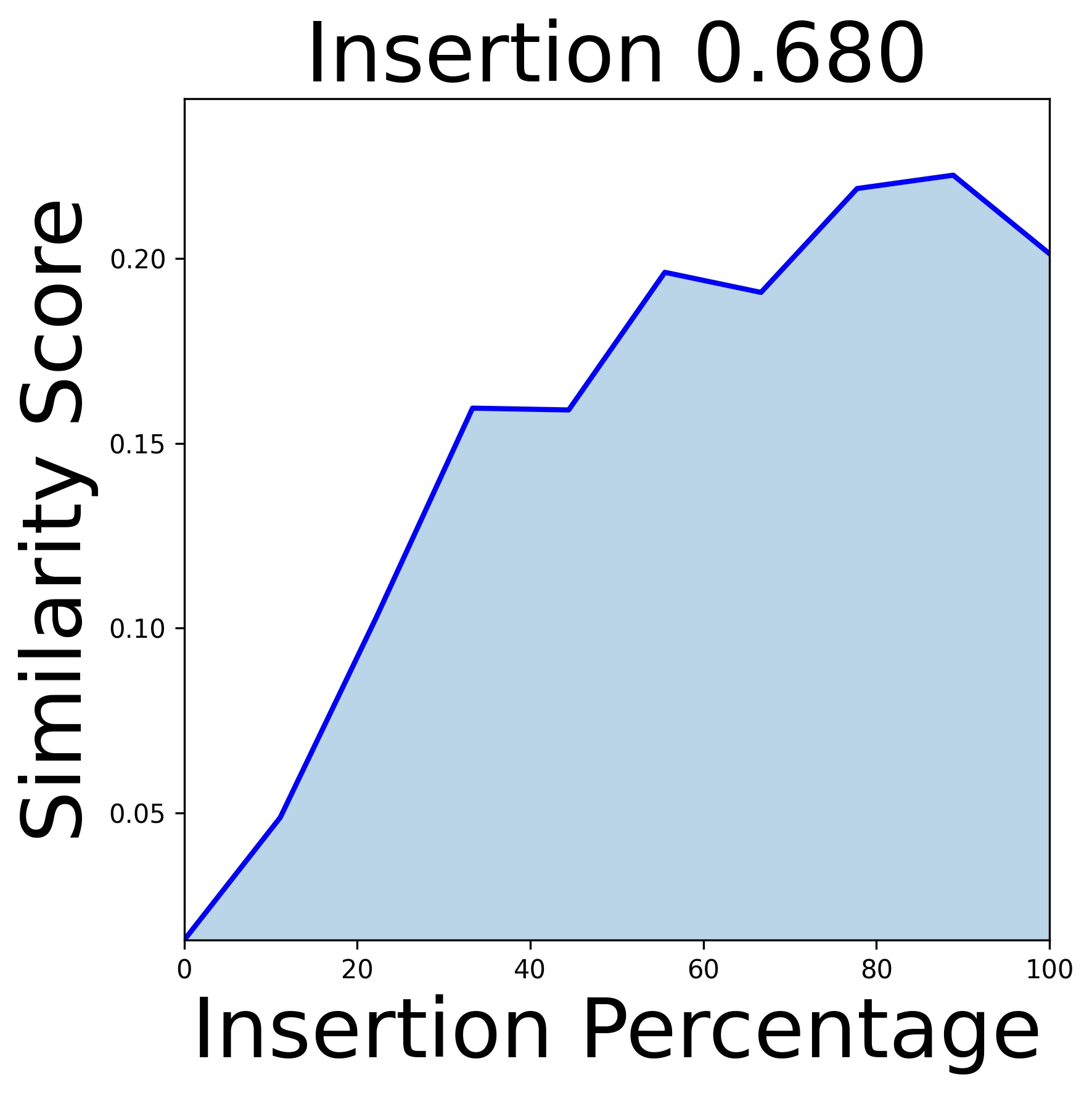}
        \includegraphics[width=0.13\textwidth]{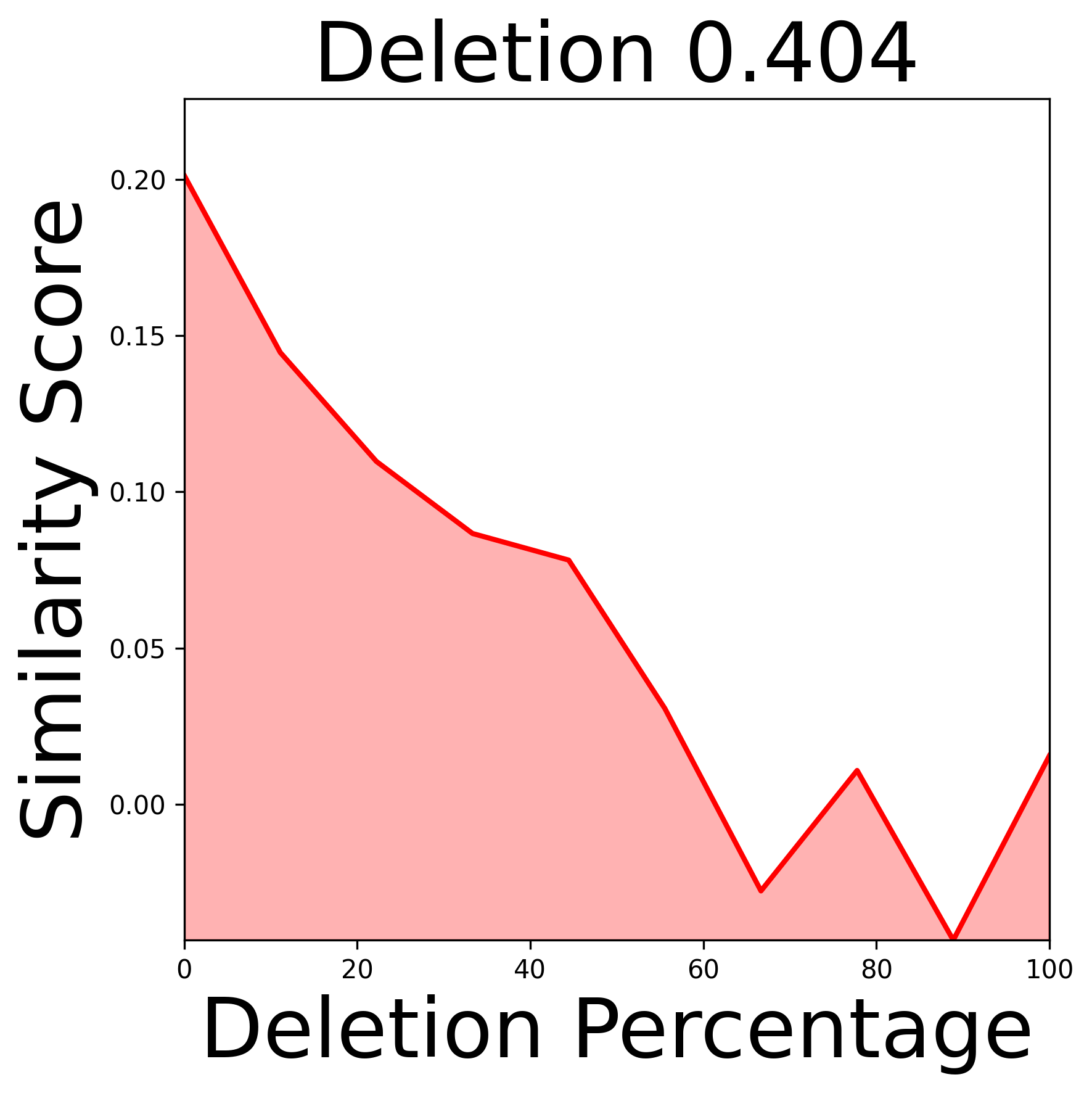}
    \end{minipage}
    
    \vspace{0.5em}

    \begin{minipage}{1\textwidth}
        \centering
        \includegraphics[width=0.13\textwidth]{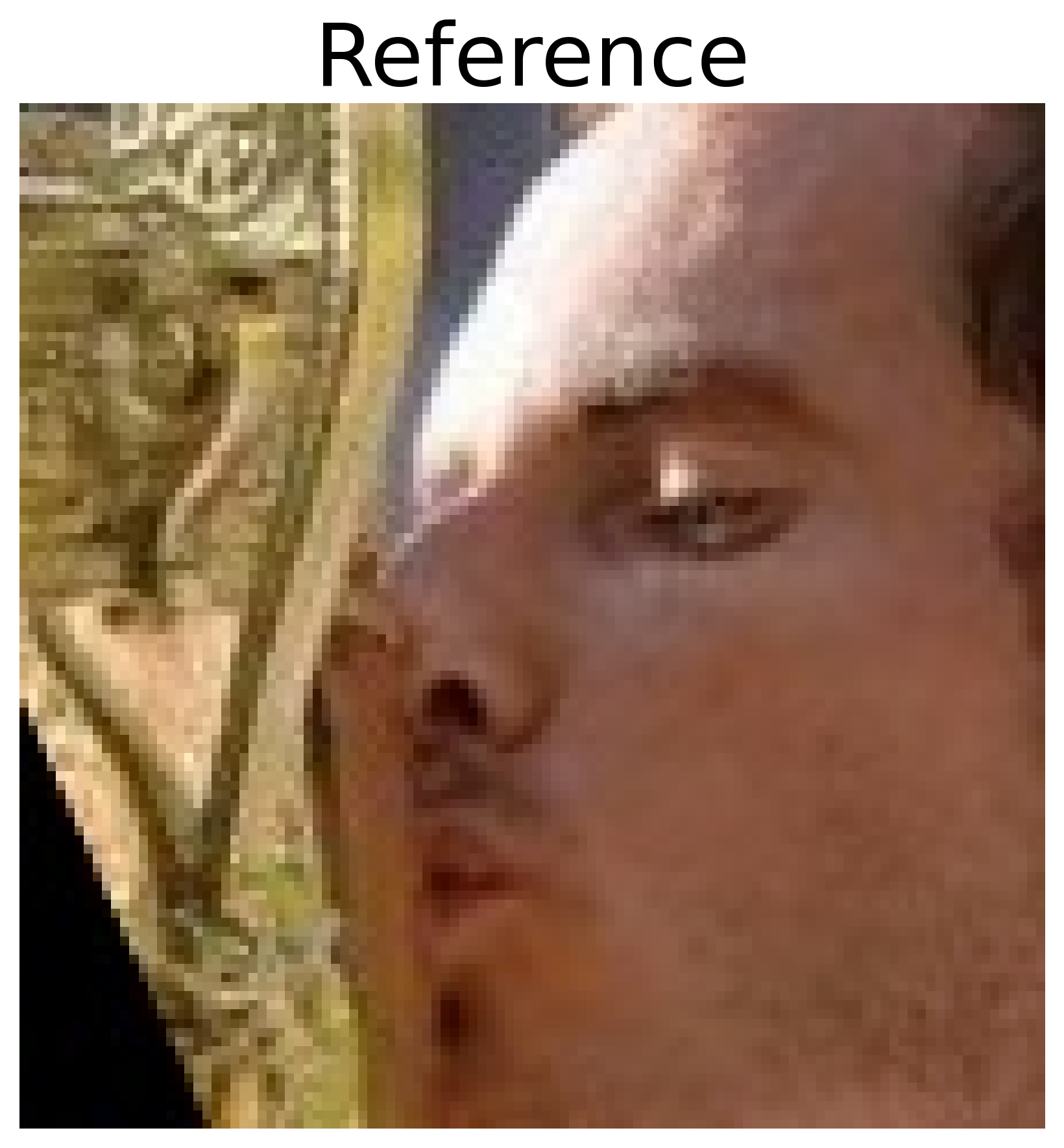}
        \raisebox{0\height}{\tikz\draw[dashed, line width=0.4pt] (0,0) -- (0,0.13\textwidth);}%
        \hspace{0.5em}
        \includegraphics[width=0.13\textwidth]{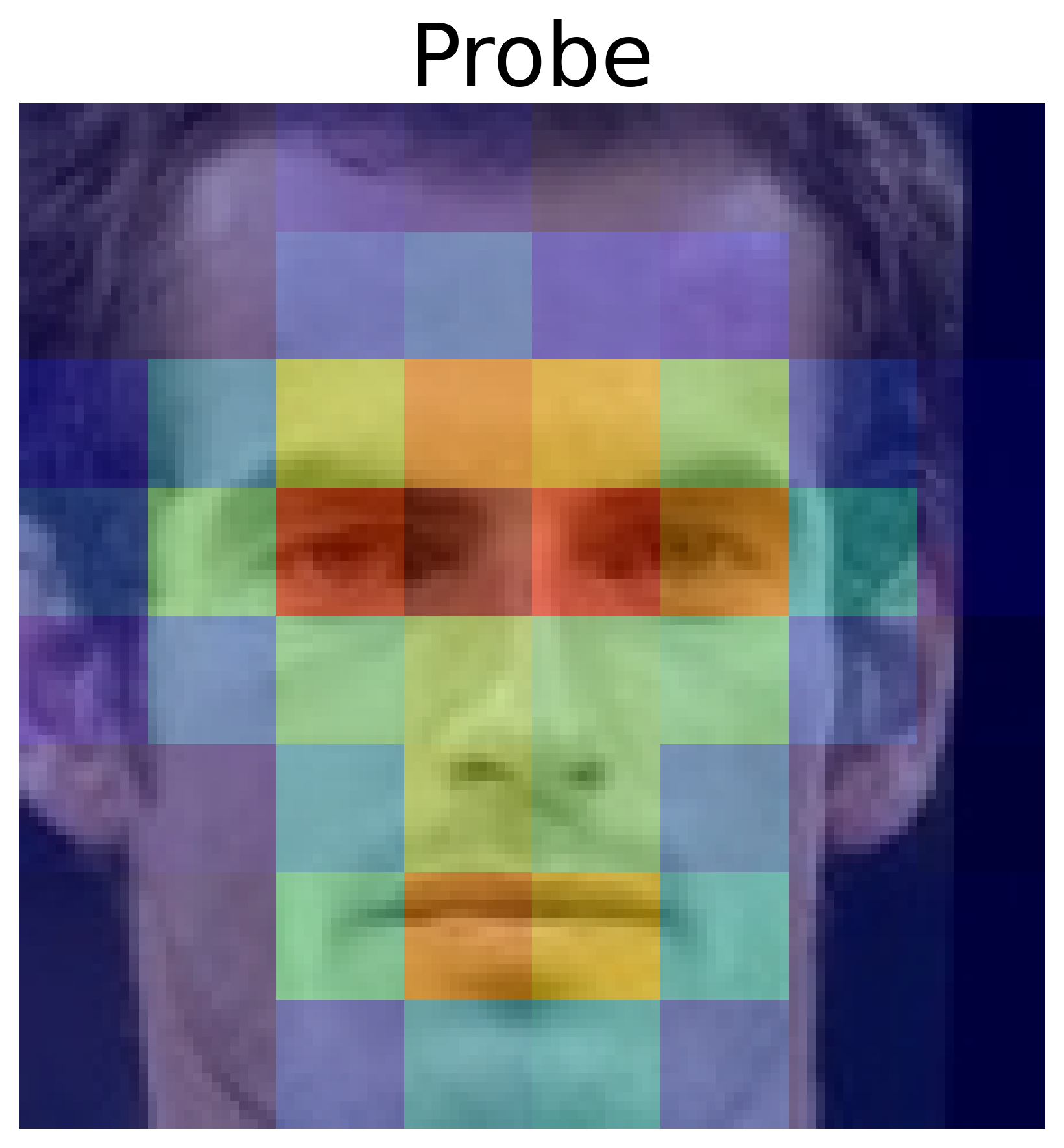}
        \includegraphics[width=0.13\textwidth]{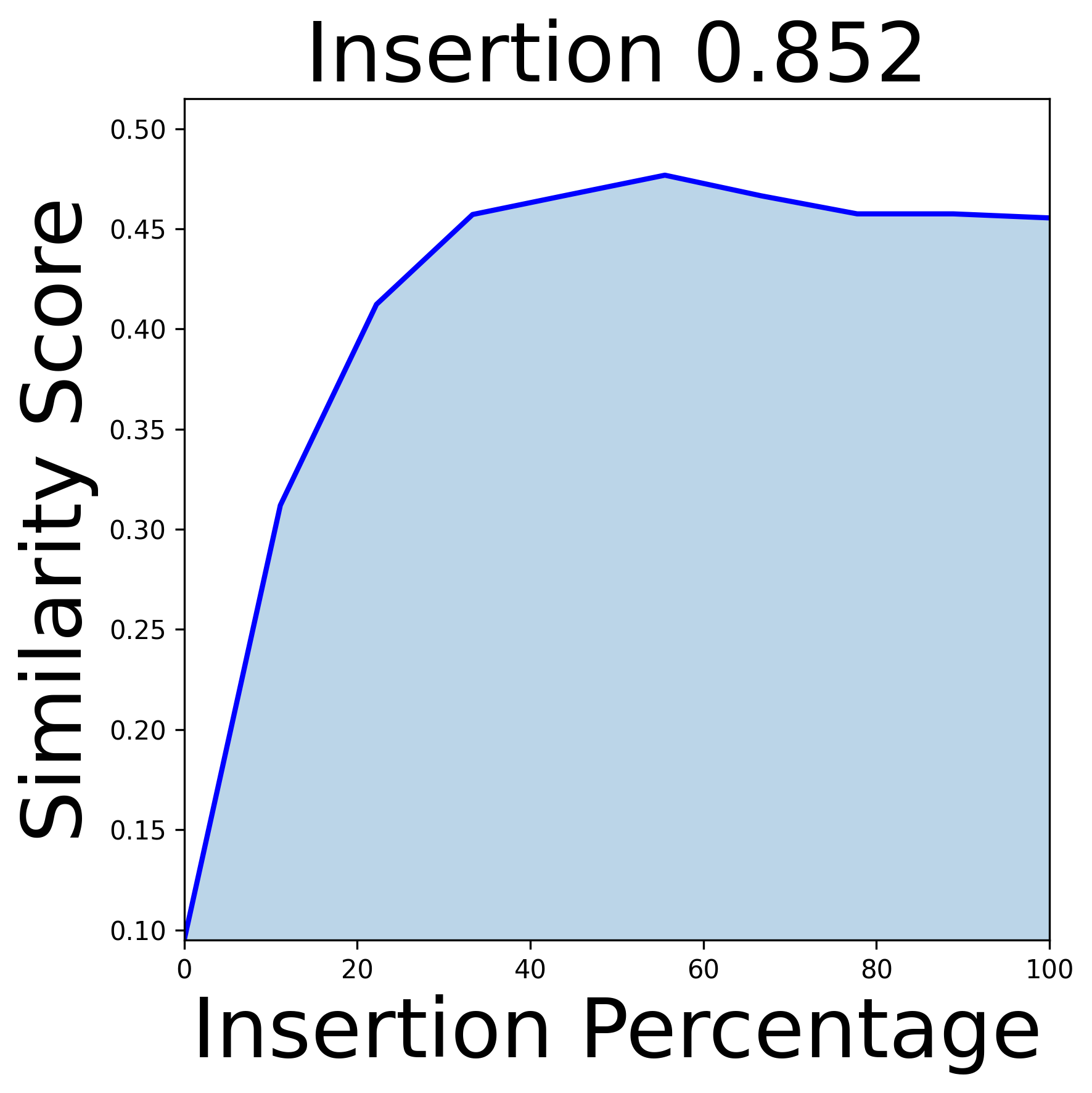}
        \includegraphics[width=0.13\textwidth]{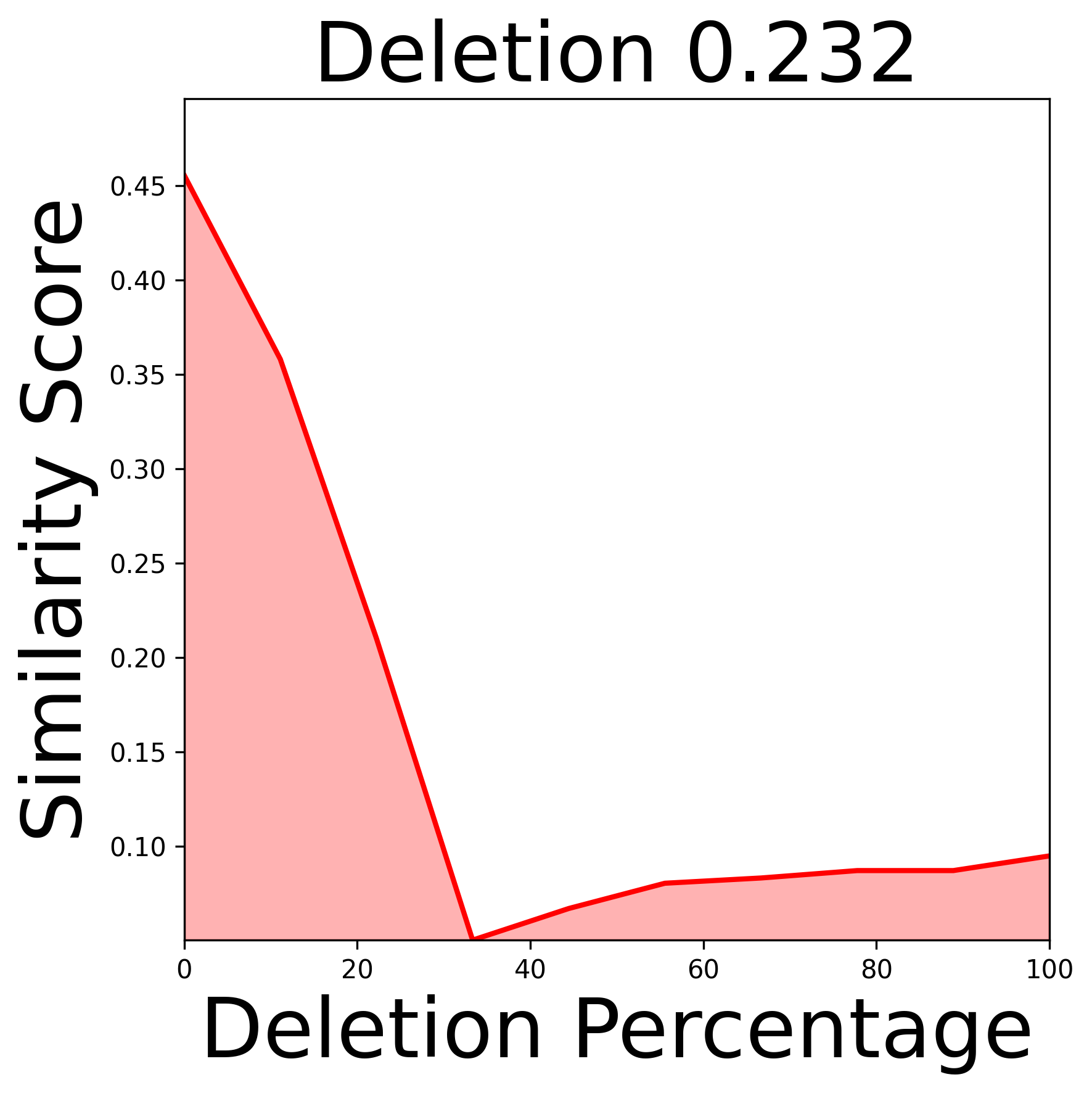}
        \raisebox{0\height}{\tikz\draw[dashed, line width=0.4pt] (0,0) -- (0,0.13\textwidth);}%
        \hspace{0.5em}
        \includegraphics[width=0.13\textwidth]{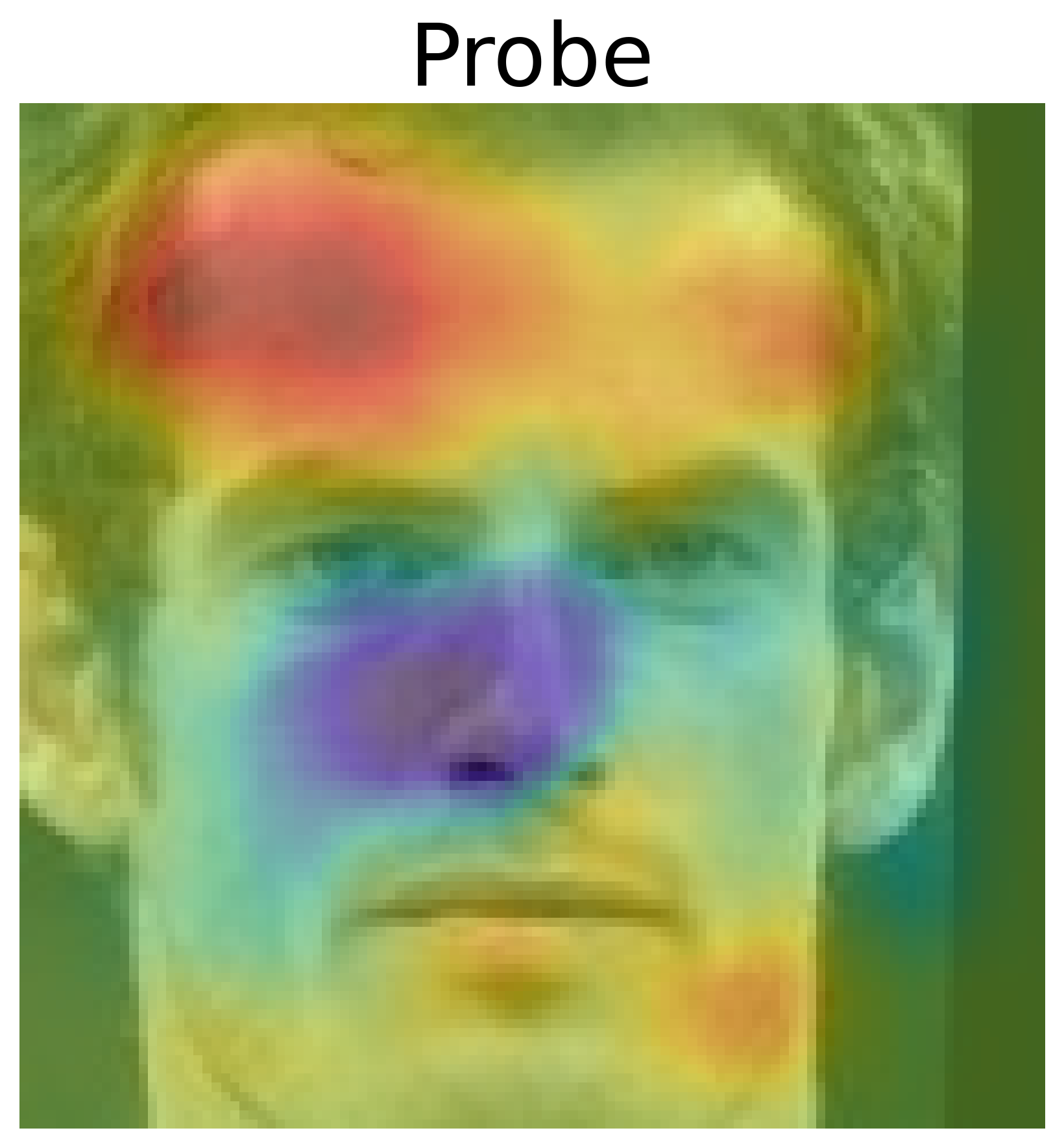}
        \includegraphics[width=0.13\textwidth]{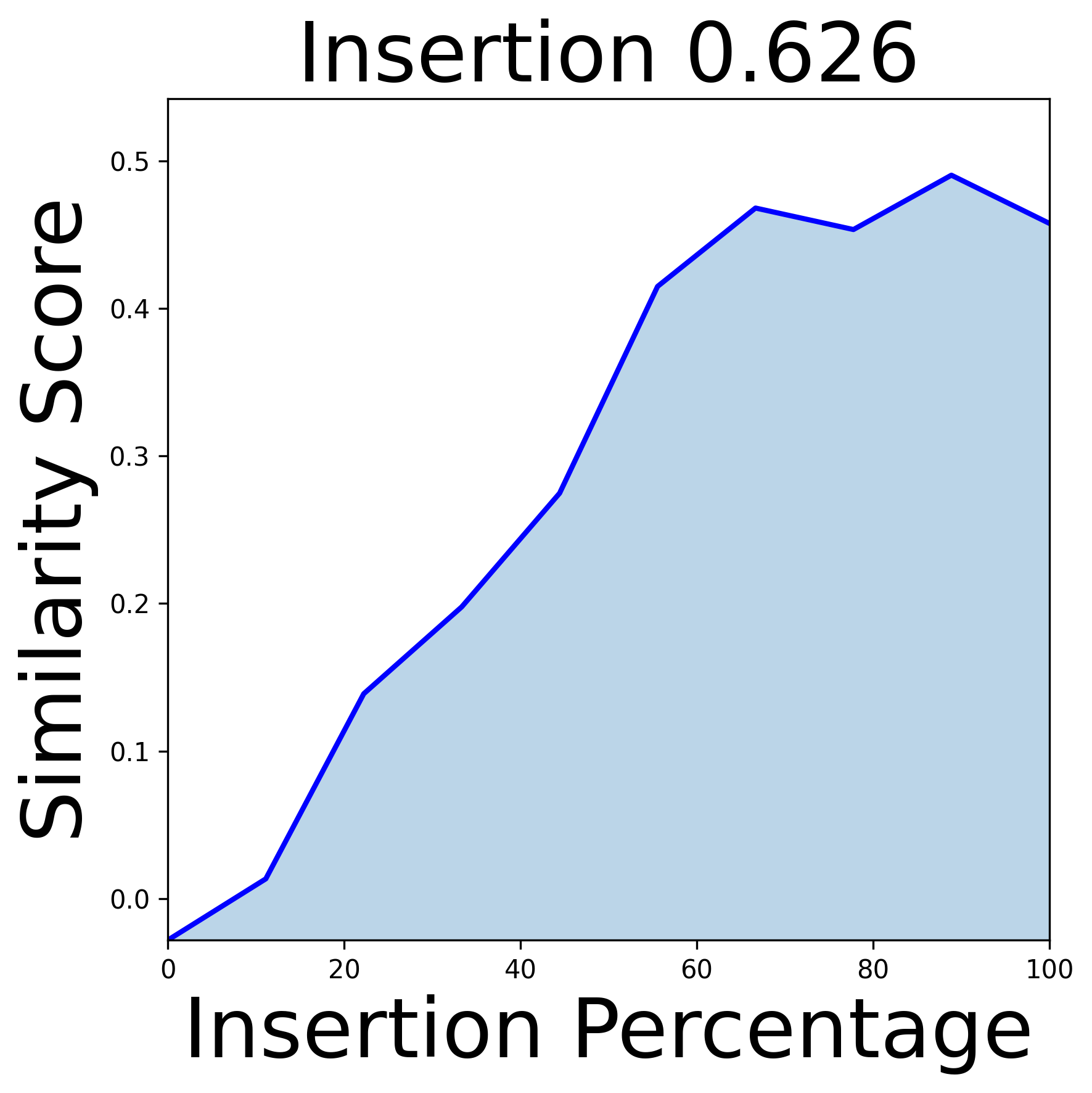}
        \includegraphics[width=0.13\textwidth]{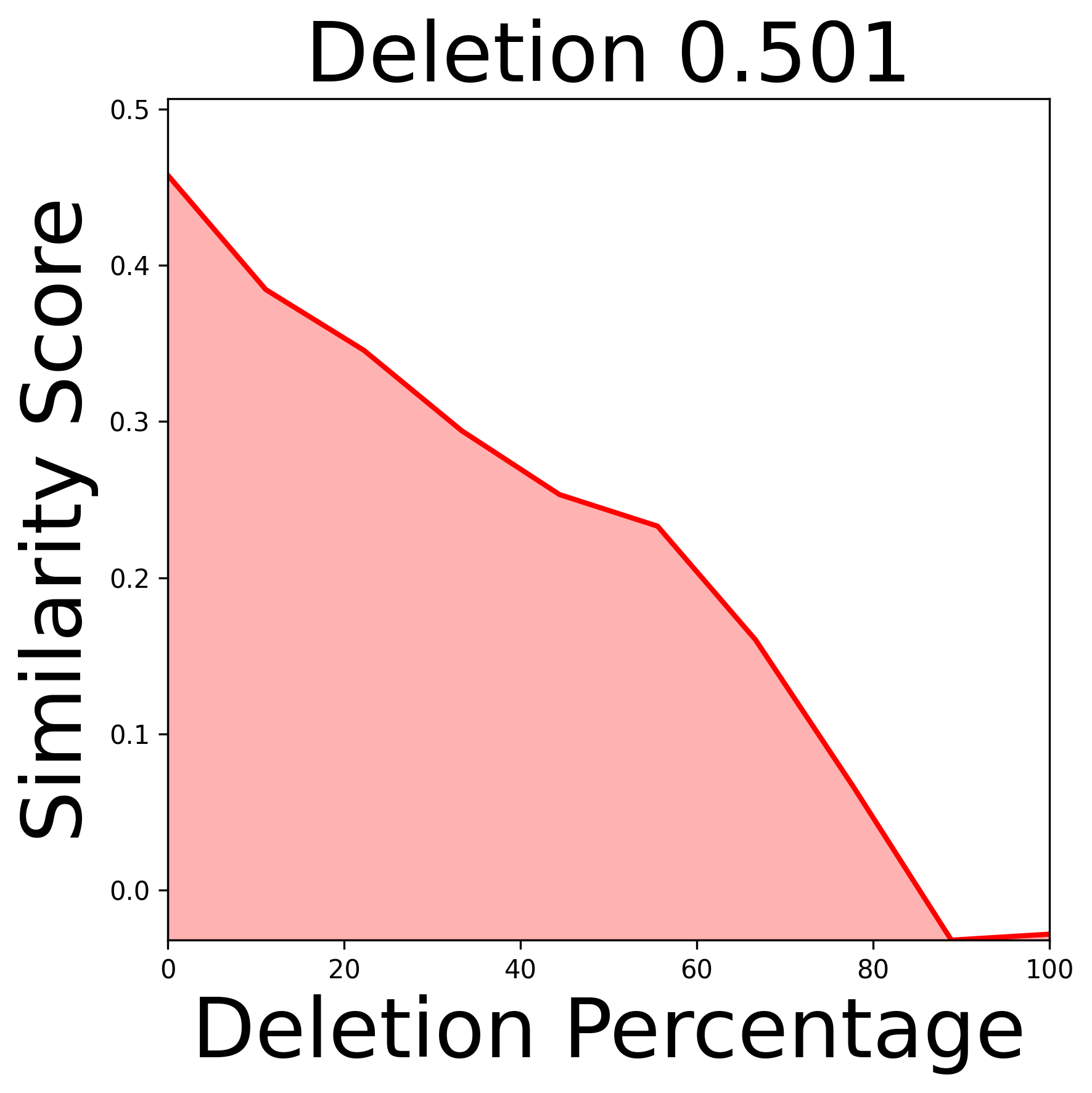}
    \end{minipage}
    
    \vspace{0.5em}
    
    \caption{Examples of genuine image pairs. For each probe image, the saliency map and corresponding insertion and deletion scores are shown for both RRFNet-28 and xFace \cite{knoche2023explainable}.}
    \label{fig:rrfnet_vs_xface}
\end{figure*}

%% file: sec/6_conclusion.tex
\section{Conclusion}
\label{sec:conclusion}

In this work, we propose RRFNet to enable a joint optimization of local feature learning and score fusion, defining global face similarity as a sum of local similarities. We demonstrate that this approach enables competitive verification rates using patches as small as small as $28 \times 28$. Moreover, we show that RRFNet-56 can achieve higher verification rates than the state-of-the-art approaches. Unlike post-hoc explanation methods that require additional processing after a decision is made, our approach improves model transparency through its inherent design, eliminating the need for extra computation to generate explanations.

In future work, we plan to investigate the automatic selection of receptive field sizes and positions to further enhance performance. Although our uniformly distributed patches across the face image yield strong results, we believe that adapting patch selection to the specific image pairs under comparison could further improve verification rates. Moreover, while we strictly follow the configurations of \cite{deng2019arcface} to ensure a fair performance comparison, we believe our approach could be further optimized using alternative CNN or ViT architectures.